\documentclass[twoside,11pt]{article}

\usepackage{blindtext}

\usepackage[abbrvbib]{jmlr2e}

\usepackage{threeparttable}  
\usepackage{enumerate}
\usepackage{graphicx}
\usepackage{float}
\usepackage{subfigure}
\usepackage{booktabs}
\usepackage{threeparttable} 
\usepackage{multirow}
\usepackage{caption}
\usepackage{lastpage}
\usepackage{natbib}
\usepackage{hyperref} 
\hypersetup{
    colorlinks=true,
    linkcolor=blue,
    filecolor=cyan,
    anchorcolor=green,
    urlcolor=blue,
    citecolor=black}

\usepackage{xcolor}
\usepackage{listings}
\lstdefinestyle{lfonts}{
  basicstyle       = \footnotesize\ttfamily,
  backgroundcolor  = \color{olive!5!white},
  stringstyle      = \color{green!60!black},
  keywordstyle     = \color{blue!90!black}\bf,  
  commentstyle     = \color{olive!70!black}\it,  
}
\lstdefinestyle{lnumbers}{
  numbers     = left,
  numberstyle = \tiny,
  numbersep   = 0.2em,
  firstnumber = 1,
  stepnumber  = 1,
}
\lstdefinestyle{llayout}{
  breaklines = true,
  tabsize    = 2,
  columns    = flexible,
}
\lstdefinestyle{lgeometry}{
  xleftmargin      = 20pt,
  xrightmargin     = 0pt,
  frame            = tb,
  framesep         = \fboxsep,
  framexleftmargin = 20pt,
}
\lstdefinestyle{lgeneral}{
  style = lfonts,
  style = lnumbers,
  style = llayout,
  style = lgeometry,
}
\lstdefinestyle{python}{
  language = {Python},
  style    = lgeneral,
}

\ShortHeadings{PyPop7: A Pure-Python Library for Population-Based Black-Box Optimization}{DUAN, ZHOU, SHAO, WANG, FENG, HUANG, TAN, YANG, ZHAO, and SHI}

\firstpageno{1}

\begin{document}

\title{PyPop7: A Pure-Python Library for Population-Based Black-Box Optimization}

\author{\name Qiqi Duan\textsuperscript{1,$\ast$} \email 11749325@mail.sustech.edu.cn
  \AND
  \name Guochen Zhou\textsuperscript{2,$\ast$} \email 12132378@mail.sustech.edu.cn
  \AND
  \name Chang Shao\textsuperscript{3,$\ast$} \email chang.shao@student.uts.edu.au
  \AND
  \name Zhuowei Wang\textsuperscript{4} \email zhuowei.wang@csiro.au
  \AND
  \name Mingyang Feng\textsuperscript{5} \email 11856010@mail.sustech.edu.cn
  \AND
  \name Yuwei Huang\textsuperscript{2} \email 12332473@mail.sustech.edu.cn
  \AND
  \name Yajing Tan\textsuperscript{2} \email 12332416@mail.sustech.edu.cn
  \AND
  \name Yijun Yang\textsuperscript{6} \email altmanyang@tencent.com
  \AND
  \name Qi Zhao\textsuperscript{2} \email zhaoq@sustech.edu.cn
  \AND
  \name Yuhui Shi\textsuperscript{2,\dag} \email shiyh@sustech.edu.cn\\
  \addr \textsuperscript{1}Harbin\ Institute\ of\ Technology,\ Harbin,\ China\\
  \textsuperscript{2}Southern\ University\ of\ Science\ and\ Technology,\ Shenzhen,\ China\\
  \textsuperscript{3}University\ of\ Technology\ Sydney,\ Sydney,\ Australia\\
  \textsuperscript{4}Space\ and\ Astronomy,\ CSIRO,\ Marshfield,\ Australia\\
  \textsuperscript{5}University\ of\ Birmingham,\ Birmingham,\ UK\\
  \textsuperscript{6}Tencent Inc.,\ Shenzhen,\ China}

\editor{}

\maketitle

{\def\thefootnote{$\ast$}\footnotetext{These three authors contributed equally.}}{}
{\def\thefootnote{\dag}\footnotetext{Corresponding author.}}{}

\begin{abstract}%
  In this paper, we present an open-source pure-Python library called \href{https://github.com/Evolutionary-Intelligence/pypop}{PyPop7} for black-box optimization (BBO). As population-based methods (e.g., evolutionary algorithms, swarm intelligence, and pattern search) become increasingly popular for BBO, the design goal of \href{https://github.com/Evolutionary-Intelligence/pypop}{PyPop7} is to provide a unified API and elegant implementations for them, particularly in challenging high-dimensional scenarios. Since these population-based methods easily suffer from the notorious curse of dimensionality owing to random sampling as one of core operations for most of them, recently various improvements and enhancements have been proposed to alleviate this issue more or less mainly via exploiting possible problem structures: such as, decomposition of search distribution or space, low-memory approximation, low-rank metric learning, variance reduction, ensemble of random subspaces, model self-adaptation, and fitness smoothing. These novel sampling strategies could better exploit different problem structures in high-dimensional search space and therefore they often result in faster rates of convergence and/or better qualities of solution for large-scale BBO. Now \href{https://github.com/Evolutionary-Intelligence/pypop}{PyPop7} has covered many of these important advances on a set of well-established BBO algorithm families and also provided an open-access interface to adding the latest or missed black-box optimizers for further functionality extensions. Its well-designed source code (under \href{https://www.gnu.org/licenses/gpl-3.0.en.html}{GPL-3.0} license) and full-fledged online documents (under \href{https://creativecommons.org/licenses/by/4.0/deed.en}{CC-BY 4.0} license) have been freely available at \url{https://github.com/Evolutionary-Intelligence/pypop} and \url{https://pypop.readthedocs.io}, respectively.
\end{abstract}


\section{Introduction}

An increasing number of population-based randomized optimization methods \citep{J_ALJ_campelo2023lessons, J_SI_aranha2022metaphorbased, J_EJOR_swan2022metaheuristics} have been widely applied to a diverse set of real-world black-box problems such as direct search \citep{C_OSDI_moritz2018ray} of deep neural network-based policy for reinforcement learning \citep{arXiv_salimans2017evolution}. In typical black-box optimization (BBO) scenarios, the lack/unavailability of gradient information severely limits the common usage of powerful gradient-based optimizers such as gradient descent \citep{J_NECO_amari1998natural} and coordinate descent \citep{J_MP_wright2015coordinate}, a problem worsened by noisy objective functions \citep{J_COA_arnold2003comparison}. Instead, a variety of black-box (aka zeroth-order or derivative-free or direct search) optimizers from multiple research communities are natural algorithm choices of practical acceptance in these challenging BBO cases \citep{C_PPSN_varelas2018comparative}. Please refer to e.g., the latest \textit{Nature} review \citep{J_NATURE_eiben2015evolutionary} or the classical \textit{Science} review \citep{J_SCIENCE_forrest1993genetic} for an introduction to population-based (also called evolution/swarm-based) optimization methods e.g., evolutionary algorithms \citep{J_NATMI_miikkulainen2021biological, B_back1997handbook}, swarm intelligence \citep{B_kennedy2001swarm, B_bonabeau1999swarm}, and pattern search \citep{J_SIOPT_torczon1997convergence}.

Over the past ten years, rapid developments of deep models \citep{J_NATURE_lecun2015deep, J_NN_schmidhuber2015deep} and big data have generated a large number of new challenging high-dimensional BBO problems, e.g., direct policy search of deep reinforcement learning \citep{arXiv_salimans2017evolution, C_OSDI_moritz2018ray}, black-box attacks of deep neural networks \citep{C_ICML_ilyas2018blackbox}, black-box prompt tuning of large language models \citep{C_ICML_sun2022blackbox}, and black-box optimization of complex generative models \citep{bioRxiv_choudhury2023generative}. These new large-scale BBO problems have greatly urged plenty of researchers from different science and engineering fields to scale up previous black-box optimizers via efficient improvements to existing (mostly random) sampling strategies \citep{C_PPSN_varelas2018comparative} or to propose novel versions of black-box optimizers targeted for large-scale scenarios, given the fact that random sampling strategies adopted by most of them suffer easily from the notorious curse of dimensionality \citep{J_FoCM_nesterov2017random, B_bellman1961adaptive}.

\begin{figure}
  \centering
  \includegraphics[width=0.98\textwidth]{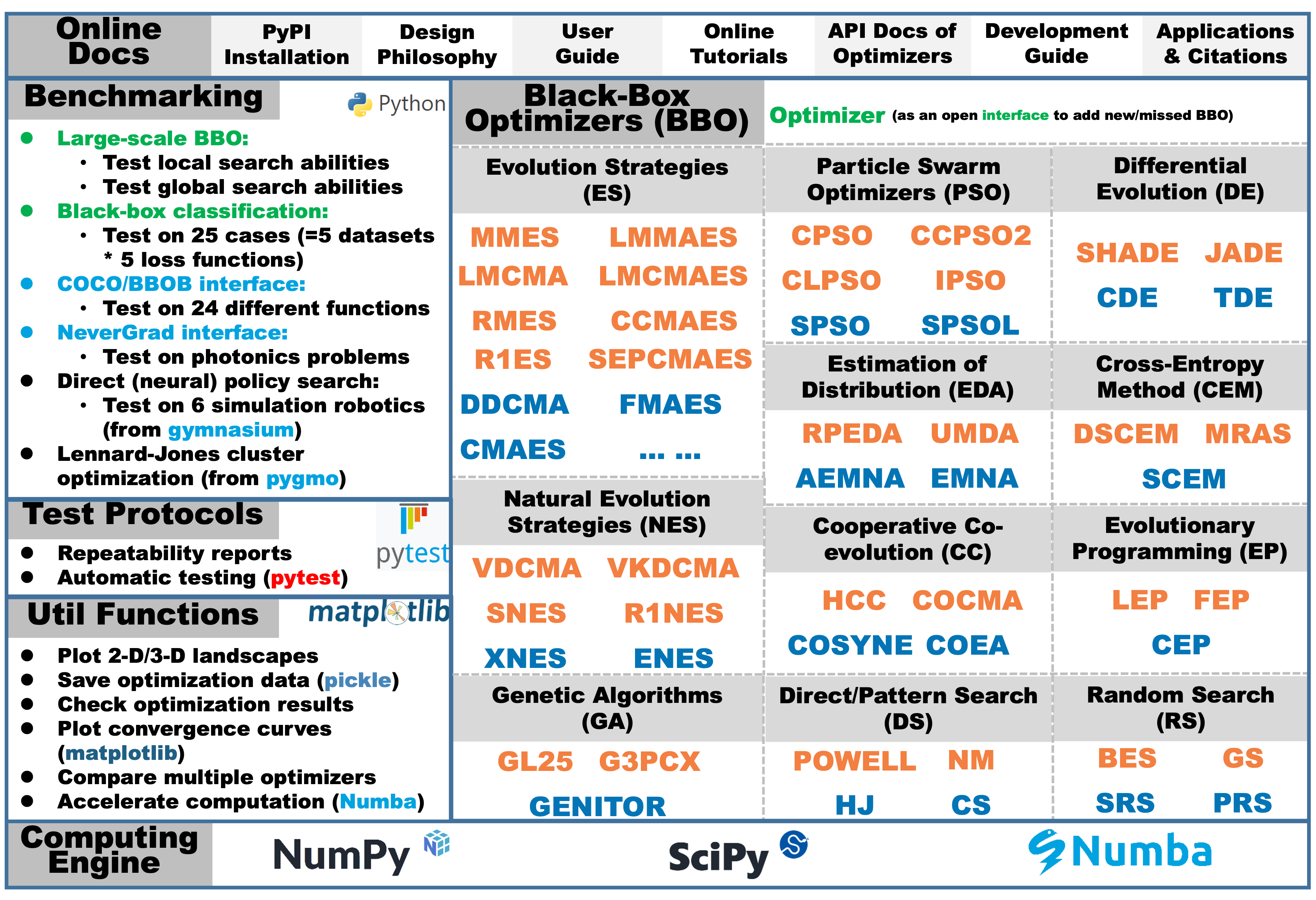}
  \caption{A conceptual framework of PyPop7 for black-box optimization (BBO), where all optimizers colored in {\color{orange}\textbf{orange}} are mainly designed for large-scale BBO.}
  \label{fig:framework}
\end{figure}

In this paper, we design an open-source pure-Python software library called \href{https://github.com/Evolutionary-Intelligence/pypop}{PyPop7}, in order to cover a large number of population-based black-box optimizers, especially their large-scale variants/versions owing to their practical potential for BBO problems of interest. Specifically, our goal is to provide a unified (API) interface and elegant implementations for them, in order to promote research repeatability \citep{J_JMLR_sonnenburg2007need}, systematic benchmarking of BBO \citep{J_OMS_hansen2021coco, J_TEVC_meunier2022blackbox}, and most importantly their \href{https://github.com/Evolutionary-Intelligence/DistributedEvolutionaryComputation}{real-world applications}. By product, we have also provided an open-access (API) interface to add the latest or missed black-box optimizers as further functionality extensions of this open-source library. Please refer to Figure~\ref{fig:framework} for its core conceptual framework, which is mainly consisting of 6 basic components (computing engines, a family of black-box optimizers, a set of util functions, two test protocols, a series of benchmarking, and full-fledged documentations). For details to each of them, please refer to Section 3 and Section 4 or its open-source repository (available at \href{https://github.com/Evolutionary-Intelligence/pypop}{GitHub}) and its online documentations (available at \href{https://pypop.readthedocs.io/en/latest/}{readthedocs.io}).

\section{Related Work}

In this section, we will introduce some existing Python libraries including population-based optimizers for BBO (e.g., evolutionary algorithms and swarm intelligence) and compare/highlight main differences between our work and them, as presented below.

Recently, \citet{J_OMS_hansen2021coco} released a well-documented benchmarking platform called \href{https://github.com/numbbo/coco}{COCO} for comparing continuous black-box optimizers, after experiencing more than 10-years developments. \href{https://github.com/numbbo/coco}{COCO}, however, focuses on the systematic design of benchmarking functions and does not provide any optimization algorithms up to now. Another similar work is the popular \href{https://github.com/facebookresearch/nevergrad}{NeverGrad} platform from Facebook Research, which covers a relatively limited number of large-scale algorithm versions \citep{D_rapin2018nevergrad}. Therefore, our pure-Python library, \href{https://github.com/Evolutionary-Intelligence/pypop}{PyPop7}, can be seen as their complement particularly for large-scale BBO. In our online \href{https://pypop.readthedocs.io/en/latest/}{tutorials}, we have shown how to connect black-box optimizers from our library with these two well-designed benchmarking platforms for BBO.

In the past, \href{https://github.com/DEAP/deap}{DEAP} \citep{J_JMLR_fortin2012deap} provided a Python platform for rapid prototyping of population-based optimizers, but leaves the challenging performance-tuning task to the end-users. This is obviously different from our library wherein performance-tuning is attributed to the developers except the coding of the fitness function to be optimized. Although \href{https://github.com/pybrain/pybrain}{PyBrain} \citep{J_JMLR_schaul2010pybrain} mainly provided a class of natural evolution strategies (NES), now it seems to be not maintained anymore and does not cover many other BBO versions in our library. The well-designed \href{https://github.com/esa/pagmo}{PaGMO} \citep{J_JOSS_biscani2020parallel} library for parallel population-based optimizers has been actively maintained for more than 10 years. However, its current focus turns to multi-objective optimization rather than large-scale BBO, which is the focus of our paper.

Overall, our Python library (called \href{https://github.com/Evolutionary-Intelligence/pypop}{PyPop7}) have provided a large set of rich and powerful optimizers for BBO from multiple research communities (e.g., artificial intelligence, machine learning, evolutionary computation, meta-heuristics, swarm intelligence, operations research, mathematical optimization, statistics, automatic control, and etc.).

\section{A Modular Coding Framework of PyPop7}

In this section, we will introduce the unified interface of \href{https://github.com/Evolutionary-Intelligence/pypop}{PyPop7} (via objective-oriented programming), testing protocols for \href{https://docs.pytest.org/}{pytest}-based automatic checking and artificially-designed repeatability reporting, its computational efficiency (via comparing \href{https://github.com/Evolutionary-Intelligence/pypop}{PyPop7} with one popular counterpart), and benchmarking on modern ML tasks for large-scale BBO.

\subsection{A Unified API for Black-Box Optimizers}

For simplicity, extensibility, and maintainability (arguably three desirable properties for any software), \href{https://github.com/Evolutionary-Intelligence/pypop}{PyPop7} has provided a unified API for a large set of black-box optimizer versions/variants within the modular coding structures based on powerful objective-oriented programming (OOP) \citep{B_lutz2013learning}. At first glance, its main organization framework is briefly summarized in Figure~\ref{fig:framework}, wherein two levels of inheritance are employed via OOP for any instantiated optimizers in order to maximize reuse and unify the design of APIs. For computational efficiency (crucial for large-scale BBO), our library depends mainly on four open-source high-performance scientific/numeric computing libraries: \href{https://numpy.org/}{\textit{NumPy}} \citep{J_NATURE_harris2020array}, \href{https://scipy.org/}{\textit{SciPy}} \citep{J_NMETH_virtanen2020scipy}, \href{https://scikit-learn.org/stable/}{\textit{Scikit-Learn}} \citep{J_JMLR_pedregosa2011scikitlearn} and \href{https://numba.pydata.org/}{\textit{Numba}} as underlying computing engines.

In our library, currently all of these black-box optimizers have been roughly classified into a total of 13 optimization algorithm classes, as presented below. To gain insights into their application cases, we have built an \href{https://github.com/Evolutionary-Intelligence/DistributedEvolutionaryComputation}{online website} to specifically collect their applications, which have been published on many (though not all) top-tier journals and conferences (such as, \href{https://www.nature.com/}{Nature}, \href{https://www.science.org/journal/science}{Science}, \href{https://www.pnas.org/}{PNAS}, \href{https://journals.aps.org/prl/}{PRL}, \href{https://pubs.acs.org/journal/jacsat}{JACS}, \href{https://proceedingsoftheieee.ieee.org/}{PIEEE}, \href{https://www.cell.com/cell/home}{Cell}, \href{https://www.jmlr.org/}{JMLR}, etc.).

\begin{itemize}
  \item Evolution Strategies: \href{https://github.com/Evolutionary-Intelligence/pypop/tree/main/pypop7/optimizers/es}{ES} \citep{J_SIOPT_akimoto2022global, C_ICML_vicol2021unbiased, J_JMLR_ollivier2017informationgeometric, J_MP_diouane2015globally, B_back2013contemporary, BS_rudolph2012evolutionary, J_NACO_beyer2002evolution, J_ECJ_hansen2001completely, J_AOOR_schwefel1984evolution, C_ISS_rechenberg1984evolution},
  \item Natural Evolution Strategies: \href{https://github.com/Evolutionary-Intelligence/pypop/tree/main/pypop7/optimizers/nes}{NES} \citep{J_JMLR_huttenrauch2024robust, J_IJCV_wei2022sparse, J_JMLR_wierstra2014natural, C_ICML_yi2009stochastic, C_CEC_wierstra2008natural},
  \item Estimation of Distribution Algorithms: \href{https://github.com/Evolutionary-Intelligence/pypop/tree/main/pypop7/optimizers/eda}{EDA} \citep{J_JMLR_zheng2023understanding, C_GECCOC_brookes2020view, B_larranaga2002estimation, C_NeurIPS_baluja1996genetic, C_ICML_baluja1995removing},
  \item Cross-Entropy Methods: \href{https://github.com/Evolutionary-Intelligence/pypop/tree/main/pypop7/optimizers/cem}{CEM} \citep{C_ICLR_wang2020exploring, C_ICML_amos2020differentiable, J_OR_hu2007model, B_rubinstein2004crossentropy, C_ICML_mannor2003cross},
  \item Differential Evolution: \href{https://github.com/Evolutionary-Intelligence/pypop/tree/main/pypop7/optimizers/de}{DE} \citep{J_APAREV_koob2023response, J_SCIENCE_higgins2023limited, J_SCIENCE_li2022attosecond, J_NATURE_laganowsky2014membrane, J_JGO_storn1997differential},
  \item Particle Swarm Optimizers: \href{https://github.com/Evolutionary-Intelligence/pypop/tree/main/pypop7/optimizers/pso}{PSO} \citep{J_NATURE_melis2024machine, J_MP_bungert2024polarized, J_SICON_huang2024consensusbased, J_MP_bolte2024swarm, J_SIMA_cipriani2022zeroinertia,J_JMLR_fornasier2021consensusbased,J_TPAMI_tang2019opening, C_ICNN_kennedy1995particle},
  \item Cooperative Coevolution: \href{https://github.com/Evolutionary-Intelligence/pypop/tree/main/pypop7/optimizers/cc}{CC} \citep{J_JMLR_gomez2008accelerated, J_JMLR_panait2008theoretical, J_NECO_schmidhuber2007training, C_ICML_fan2003utilizing, J_ECJ_potter2000cooperative, C_IJCAI_gomez1999solving, J_ML_moriarty1996efficient, C_ICML_moriarty1995efficient, C_PPSN_potter1994cooperative},
  \item Simulated Annealing\footnote{Note that \href{https://github.com/Evolutionary-Intelligence/pypop/tree/main/pypop7/optimizers/sa}{SA} is an individual-based rather than population-based optimization method.}: \href{https://github.com/Evolutionary-Intelligence/pypop/tree/main/pypop7/optimizers/sa}{SA} \citep{J_BIOMET_samyak2024statistical, J_JMLR_bouttier2019convergence, J_TOMS_siarry1997enhanced, J_STATS_bertsimas1993simulated, J_TOMS_corana1987minimizing, J_SCIENCE_kirkpatrick1983optimization, J_BIOMET_hastings1970monte, J_JCP_metropolis1953equation},
  \item Genetic Algorithms: \href{https://github.com/Evolutionary-Intelligence/pypop/tree/main/pypop7/optimizers/ga}{GA} \citep{J_NATURE_chen2020classification, BS_whitley2019next, J_CACM_goldberg1994genetic, J_SCIENCE_forrest1993genetic, C_NeurIPS_mitchell1993when, J_ML_goldberg1988genetic, J_JACM_holland1962outline},
  \item Evolutionary Programming: \href{https://github.com/Evolutionary-Intelligence/pypop/tree/main/pypop7/optimizers/ep}{EP} \citep{J_MS_cui2006machine, J_TEVC_yao1999evolutionary, BS_fogel1999overview, C_ECAE_fogel1995introduction, J_SAC_fogel1994evolutionary, J_THFE_fogel1965intelligent},
  \item Pattern/Direct Search: \href{https://github.com/Evolutionary-Intelligence/pypop/tree/main/pypop7/optimizers/ds}{PS/DS} \citep{J_SIREV_kolda2003optimization,J_SIOPT_lagarias1998convergence, BS_wright1996direct, J_COMJNL_nelder1965simplex, J_COMJNL_powell1964efficient, J_CACM_kaupe1963algorithm, J_JACM_hooke1961direct, R_fermi1952numerical},
  \item Random Search: \href{https://github.com/Evolutionary-Intelligence/pypop/tree/main/pypop7/optimizers/rs}{RS} \citep{J_FoCM_nesterov2017random, C_PPSN_stich2014low, J_JMLR_bergstra2012random, BS_schmidhuber2001evaluating, C_IJCAI_rosenstein2001robot, J_MOOR_solis1981minimization, J_TAC_schumer1968adaptive, J_ARC_rastrigin1963convergence, J_OR_brooks1958discussion}, and
  \item Bayesian Optimization: \href{https://github.com/Evolutionary-Intelligence/pypop/tree/main/pypop7/optimizers/bo}{BO} \citep{C_NeurIPS_wang2020learning, J_PIEEE_shahriari2016taking, J_JGO_jones1998efficient}.
\end{itemize}

To alleviate their curse of dimensionality \citep{B_bellman1957dynamic} for large-scale BBO, different kinds of sophisticated strategies have been employed to enhance these black-box optimizers, as presented in the following:

\begin{enumerate}[\ \ \ \ 1)]
  \item Decomposition of search distribution \citep{J_ECJ_akimoto2020diagonal, B_back2013contemporary, C_GECCO_schaul2011high, C_PPSN_ros2008simple} or search space \citep{J_JMLR_panait2008theoretical, C_GECCO_gomez2005coevolving, J_TOMS_siarry1997enhanced, J_TOMS_corana1987minimizing},
  \item Recursive spatial partitioning, e.g., via Monte Carlo tree search \citep{C_NeurIPS_wang2020learning},
  \item Low-memory approximation for covariance matrix adaptation \citep{J_TEVC_he2021mmes, J_TEVC_loshchilov2019large, J_ECJ_loshchilov2017lmcma, C_NeurIPS_krause2016cmaes},
  \item Low-rank metric learning \citep{J_TEVC_li2018simple, C_GECCOC_sun2013linear},
  \item Variance-reduction \citep{C_ICML_gao2022generalizing, C_PPSN_brockhoff2010mirrored},
  \item Ensemble of random subspaces constructed via random matrix theory \citep{J_SISC_demo2021supervised, J_ECJ_kaban2016largescale},
  \item Meta-model self-adaptation \citep{C_PPSN_akimoto2016online, J_TEVC_lee2004evolutionarya},
  \item Smoothing of fitness expectation \citep{J_JMLR_huttenrauch2024robust, C_ICML_gao2022generalizing, J_FoCM_nesterov2017random},
  \item Smoothing of sampling operation \citep{J_MP_bungert2024polarized, C_ICML_amos2020differentiable, J_ECJ_deb2002computationally}, and
  \item Efficient allocation of computational resources \citep{J_EJOR_garcia-martinez2008global}.
\end{enumerate}

In this new Python library \href{https://github.com/Evolutionary-Intelligence/pypop}{PyPop7}, we aim to provide high-quality open-source implementations to many of these advanced techniques on population-based optimizers for large-scale BBO in a unified way (which have been summarized in Figure~\ref{fig:framework}).

\subsection{Testing Protocols}

Importantly, in order to ensure the coding correctness of black-box optimizers, we have provided an open-access code-based repeatability report for each black-box optimizer. Specifically, for each black-box optimizer, all experimental details are given in a specific folder (corresponding to a hyperlink in the Examples section of its online API documentation) and main results generated for it are compared to reported results in its original literature. For all optimizers with repeatability reports unavailable owing to specific reasons, their Python3-based implementations have been checked carefully by three authors (and perhaps other users) to avoid trivial bugs and errors. For any failed repeatability experiment, we try our best to reach an agreement regarding some possible reason(s), which is also finally described in its repeatability report. All repeatability code/results are summarized in Table~\ref{tab:repeatability_check}, wherein each hyperlink is used to navigate the used Python code or generated results.

Following the standard workflow practice of open-source software, we have used the popular \href{https://docs.pytest.org}{pytest} tool and the free \href{https://circleci.com/}{circleci} service to automate all light-weighted testing processes.

For any randomized black-box optimizer, properly controlling its random sampling process is very important to repeat its entire optimization experiments. In our library, the random seed for each black-box optimizer should be explicitly set in order to ensure maximal repeatability, according to the newest \href{https://numpy.org/doc/stable/reference/random/index.html}{suggestion} from \href{https://numpy.org/}{\textit{NumPy}} for random sampling.

\begin{table}[htbp]
  \centering
  \caption{\textbf{Repeatability Reports of All Black-Box Optimizers from PyPop7}}
  \label{tab:repeatability_check}
  \resizebox{\linewidth}{0.65\linewidth}{
    \begin{tabular}{c|c|c|c||c|c|c|c}
      \toprule
      \textbf{Optimizer} & \textbf{Repeatability Code}                                                                                                             & \textbf{Results}                                                                                                  & \textbf{Success} & \textbf{Optimizer} & \textbf{Repeatability Code}                                                                                                             & \textbf{Results}                                                                                                   & \textbf{Success} \\ \toprule
      MMES               & \href{https://github.com/Evolutionary-Intelligence/pypop/blob/main/pypop7/optimizers/es/_repeat_mmes.py}{\_repeat\_mmes.py}             & \href{https://github.com/Evolutionary-Intelligence/pypop/tree/main/docs/repeatability/mmes}{figures}              & YES              & FCMAES             & \href{https://github.com/Evolutionary-Intelligence/pypop/blob/main/pypop7/optimizers/es/_repeat_fcmaes.py}{\_repeat\_fcmaes.py}         & \href{https://github.com/Evolutionary-Intelligence/pypop/tree/main/docs/repeatability/fcmaes}{figures}             & YES              \\ \midrule
      LMMAES             & \href{https://github.com/Evolutionary-Intelligence/pypop/blob/main/pypop7/optimizers/es/_repeat_lmmaes.py}{\_repeat\_lmmaes.py}         & \href{https://github.com/Evolutionary-Intelligence/pypop/tree/main/docs/repeatability/lmmaes}{figures}            & YES              & LMCMA              & \href{https://github.com/Evolutionary-Intelligence/pypop/blob/main/pypop7/optimizers/es/_repeat_lmcma.py}{\_repeat\_lmcma.py}           & \href{https://github.com/Evolutionary-Intelligence/pypop/tree/main/docs/repeatability/lmcma}{figures}              & YES              \\ \midrule
      LMCMAES            & \href{https://github.com/Evolutionary-Intelligence/pypop/blob/main/pypop7/optimizers/es/_repeat_lmcmaes.py}{\_repeat\_lmcmaes.py}       & \href{https://github.com/Evolutionary-Intelligence/pypop/blob/main/pypop7/optimizers/es/_repeat_lmcmaes.py}{data} & YES              & RMES               & \href{https://github.com/Evolutionary-Intelligence/pypop/blob/main/pypop7/optimizers/es/_repeat_rmes.py}{\_repeat\_rmes.py}             & \href{https://github.com/Evolutionary-Intelligence/pypop/tree/main/docs/repeatability/rmes}{figures}               & YES              \\ \midrule
      R1ES               & \href{https://github.com/Evolutionary-Intelligence/pypop/blob/main/pypop7/optimizers/es/_repeat_r1es.py}{\_repeat\_r1es.py}             & \href{https://github.com/Evolutionary-Intelligence/pypop/tree/main/docs/repeatability/r1es}{figures}              & YES              & VKDCMA             & \href{https://github.com/Evolutionary-Intelligence/pypop/blob/main/pypop7/optimizers/es/_repeat_vkdcma.py}{\_repeat\_vkdcma.py}         & \href{https://github.com/Evolutionary-Intelligence/pypop/blob/main/pypop7/optimizers/es/_repeat_vkdcma.py}{data}   & YES              \\ \midrule
      VDCMA              & \href{https://github.com/Evolutionary-Intelligence/pypop/blob/main/pypop7/optimizers/es/_repeat_vdcma.py}{\_repeat\_vdcma.py}           & \href{https://github.com/Evolutionary-Intelligence/pypop/blob/main/pypop7/optimizers/es/_repeat_vdcma.py}{data}   & YES              & CCMAES2016         & \href{https://github.com/Evolutionary-Intelligence/pypop/blob/main/pypop7/optimizers/es/_repeat_ccmaes2016.py}{\_repeat\_ccmaes2016.py} & \href{https://github.com/Evolutionary-Intelligence/pypop/tree/main/docs/repeatability/ccmaes2016}{figures}         & YES              \\ \midrule
      OPOA2015           & \href{https://github.com/Evolutionary-Intelligence/pypop/blob/main/pypop7/optimizers/es/_repeat_opoa2015.py}{\_repeat\_opoa2015.py}     & \href{https://github.com/Evolutionary-Intelligence/pypop/tree/main/docs/repeatability/opoa2015}{figures}          & YES              & OPOA2010           & \href{https://github.com/Evolutionary-Intelligence/pypop/blob/main/pypop7/optimizers/es/_repeat_opoa2010.py}{\_repeat\_opoa2010.py}     & \href{https://github.com/Evolutionary-Intelligence/pypop/tree/main/docs/repeatability/opoa2010}{figures}           & YES              \\ \midrule
      CCMAES2009         & \href{https://github.com/Evolutionary-Intelligence/pypop/blob/main/pypop7/optimizers/es/_repeat_ccmaes2009.py}{\_repeat\_ccmaes2009.py} & \href{https://github.com/Evolutionary-Intelligence/pypop/tree/main/docs/repeatability/ccmaes2009}{figures}        & YES              & OPOC2009           & \href{https://github.com/Evolutionary-Intelligence/pypop/blob/main/pypop7/optimizers/es/_repeat_opoc2009.py}{\_repeat\_opoc2009.py}     & \href{https://github.com/Evolutionary-Intelligence/pypop/tree/main/docs/repeatability/opoc2009}{figures}           & YES              \\ \midrule
      OPOC2006           & \href{https://github.com/Evolutionary-Intelligence/pypop/blob/main/pypop7/optimizers/es/_repeat_opoc2006.py}{\_repeat\_opoc2006.py}     & \href{https://github.com/Evolutionary-Intelligence/pypop/tree/main/docs/repeatability/opoc2006}{figures}          & YES              & SEPCMAES           & \href{https://github.com/Evolutionary-Intelligence/pypop/blob/main/pypop7/optimizers/es/_repeat_sepcmaes.py}{\_repeat\_sepcmaes.py}     & \href{https://github.com/Evolutionary-Intelligence/pypop/blob/main/pypop7/optimizers/es/_repeat_sepcmaes.py}{data} & YES              \\ \midrule
      DDCMA              & \href{https://github.com/Evolutionary-Intelligence/pypop/blob/main/pypop7/optimizers/es/_repeat_ddcma.py}{\_repeat\_ddcma.py}           & \href{https://github.com/Evolutionary-Intelligence/pypop/blob/main/pypop7/optimizers/es/_repeat_ddcma.py}{data}   & YES              & MAES               & \href{https://github.com/Evolutionary-Intelligence/pypop/blob/main/pypop7/optimizers/es/_repeat_maes.py}{\_repeat\_maes.py}             & \href{https://github.com/Evolutionary-Intelligence/pypop/tree/main/docs/repeatability/maes}{figures}               & YES              \\ \midrule
      FMAES              & \href{https://github.com/Evolutionary-Intelligence/pypop/blob/main/pypop7/optimizers/es/_repeat_fmaes.py}{\_repeat\_fmaes.py}           & \href{https://github.com/Evolutionary-Intelligence/pypop/tree/main/docs/repeatability/fmaes}{figures}             & YES              & CMAES              & \href{https://github.com/Evolutionary-Intelligence/pypop/blob/main/pypop7/optimizers/es/_repeat_cmaes.py}{\_repeat\_cmaes.py}           & \href{https://github.com/Evolutionary-Intelligence/pypop/blob/main/pypop7/optimizers/es/_repeat_cmaes.py}{data}    & YES              \\ \midrule
      SAMAES             & \href{https://github.com/Evolutionary-Intelligence/pypop/blob/main/pypop7/optimizers/es/_repeat_samaes.py}{\_repeat\_samaes.py}         & \href{https://github.com/Evolutionary-Intelligence/pypop/tree/main/docs/repeatability/samaes}{figure}             & YES              & SAES               & \href{https://github.com/Evolutionary-Intelligence/pypop/blob/main/pypop7/optimizers/es/_repeat_saes.py}{\_repeat\_saes.py}             & \href{https://github.com/Evolutionary-Intelligence/pypop/blob/main/pypop7/optimizers/es/_repeat_saes.py}{data}     & YES              \\ \midrule
      CSAES              & \href{https://github.com/Evolutionary-Intelligence/pypop/blob/main/pypop7/optimizers/es/_repeat_csaes.py}{\_repeat\_csaes.py}           & \href{https://github.com/Evolutionary-Intelligence/pypop/tree/main/docs/repeatability/csaes}{figure}              & YES              & DSAES              & \href{https://github.com/Evolutionary-Intelligence/pypop/blob/main/pypop7/optimizers/es/_repeat_dsaes.py}{\_repeat\_dsaes.py}           & \href{https://github.com/Evolutionary-Intelligence/pypop/tree/main/docs/repeatability/dsaes}{figure}               & YES              \\ \midrule
      SSAES              & \href{https://github.com/Evolutionary-Intelligence/pypop/blob/main/pypop7/optimizers/es/_repeat_ssaes.py}{\_repeat\_ssaes.py}           & \href{https://github.com/Evolutionary-Intelligence/pypop/tree/main/docs/repeatability/ssaes}{figure}              & YES              & RES                & \href{https://github.com/Evolutionary-Intelligence/pypop/blob/main/pypop7/optimizers/es/_repeat_res.py}{\_repeat\_res.py}               & \href{https://github.com/Evolutionary-Intelligence/pypop/tree/main/docs/repeatability/res}{figure}                 & YES              \\ \midrule
      R1NES              & \href{https://github.com/Evolutionary-Intelligence/pypop/blob/main/pypop7/optimizers/nes/_repeat_r1nes.py}{\_repeat\_r1nes.py}          & \href{https://github.com/Evolutionary-Intelligence/pypop/blob/main/pypop7/optimizers/nes/_repeat_r1nes.py}{data}  & YES              & SNES               & \href{https://github.com/Evolutionary-Intelligence/pypop/blob/main/pypop7/optimizers/nes/_repeat_snes.py}{\_repeat\_snes.py}            & \href{https://github.com/Evolutionary-Intelligence/pypop/blob/main/pypop7/optimizers/nes/_repeat_snes.py}{data}    & YES              \\ \midrule
      XNES               & \href{https://github.com/Evolutionary-Intelligence/pypop/blob/main/pypop7/optimizers/nes/_repeat_xnes.py}{\_repeat\_xnes.py}            & \href{https://github.com/Evolutionary-Intelligence/pypop/blob/main/pypop7/optimizers/nes/_repeat_xnes.py}{data}   & YES              & ENES               & \href{https://github.com/Evolutionary-Intelligence/pypop/blob/main/pypop7/optimizers/nes/_repeat_enes.py}{\_repeat\_enes.py}            & \href{https://github.com/Evolutionary-Intelligence/pypop/blob/main/pypop7/optimizers/nes/_repeat_enes.py}{data}    & YES              \\ \midrule
      ONES               & \href{https://github.com/Evolutionary-Intelligence/pypop/blob/main/pypop7/optimizers/nes/_repeat_ones.py}{\_repeat\_ones.py}            & \href{https://github.com/Evolutionary-Intelligence/pypop/blob/main/pypop7/optimizers/nes/_repeat_ones.py}{data}   & YES              & SGES               & \href{https://github.com/Evolutionary-Intelligence/pypop/blob/main/pypop7/optimizers/nes/_repeat_sges.py}{\_repeat\_sges.py}            & \href{https://github.com/Evolutionary-Intelligence/pypop/blob/main/pypop7/optimizers/nes/_repeat_sges.py}{data}    & YES              \\ \midrule
      RPEDA              & \href{https://github.com/Evolutionary-Intelligence/pypop/blob/main/pypop7/optimizers/eda/_repeat_rpeda.py}{\_repeat\_rpeda.py}          & \href{https://github.com/Evolutionary-Intelligence/pypop/blob/main/pypop7/optimizers/eda/_repeat_rpeda.py}{data}  & YES              & UMDA               & \href{https://github.com/Evolutionary-Intelligence/pypop/blob/main/pypop7/optimizers/eda/_repeat_umda.py}{\_repeat\_umda.py}            & \href{https://github.com/Evolutionary-Intelligence/pypop/blob/main/pypop7/optimizers/eda/_repeat_umda.py}{data}    & YES              \\ \midrule
      AEMNA              & \href{https://github.com/Evolutionary-Intelligence/pypop/blob/main/pypop7/optimizers/eda/_repeat_aemna.py}{\_repeat\_aemna.py}          & \href{https://github.com/Evolutionary-Intelligence/pypop/blob/main/pypop7/optimizers/eda/_repeat_aemna.py}{data}  & YES              & EMNA               & \href{https://github.com/Evolutionary-Intelligence/pypop/blob/main/pypop7/optimizers/eda/_repeat_emna.py}{\_repeat\_emna.py}            & \href{https://github.com/Evolutionary-Intelligence/pypop/blob/main/pypop7/optimizers/eda/_repeat_emna.py}{data}    & YES              \\ \midrule
      DCEM               & \href{https://github.com/Evolutionary-Intelligence/pypop/blob/main/pypop7/optimizers/cem/_repeat_dcem.py}{\_repeat\_dcem.py}            & \href{https://github.com/Evolutionary-Intelligence/pypop/blob/main/pypop7/optimizers/cem/_repeat_dcem.py}{data}   & YES              & DSCEM              & \href{https://github.com/Evolutionary-Intelligence/pypop/blob/main/pypop7/optimizers/cem/_repeat_dscem.py}{\_repeat\_dscem.py}          & \href{https://github.com/Evolutionary-Intelligence/pypop/blob/main/pypop7/optimizers/cem/_repeat_dscem.py}{data}   & YES              \\ \midrule
      MRAS               & \href{https://github.com/Evolutionary-Intelligence/pypop/blob/main/pypop7/optimizers/cem/_repeat_mras.py}{\_repeat\_mras.py}            & \href{https://github.com/Evolutionary-Intelligence/pypop/blob/main/pypop7/optimizers/cem/_repeat_mras.py}{data}   & YES              & SCEM               & \href{https://github.com/Evolutionary-Intelligence/pypop/blob/main/pypop7/optimizers/cem/_repeat_scem.py}{\_repeat\_scem.py}            & \href{https://github.com/Evolutionary-Intelligence/pypop/blob/main/pypop7/optimizers/cem/_repeat_scem.py}{data}    & YES              \\ \midrule
      SHADE              & \href{https://github.com/Evolutionary-Intelligence/pypop/blob/main/pypop7/optimizers/de/_repeat_shade.py}{\_repeat\_shade.py}           & \href{https://github.com/Evolutionary-Intelligence/pypop/blob/main/pypop7/optimizers/de/_repeat_shade.py}{data}   & YES              & JADE               & \href{https://github.com/Evolutionary-Intelligence/pypop/blob/main/pypop7/optimizers/de/_repeat_jade.py}{\_repeat\_jade.py}             & \href{https://github.com/Evolutionary-Intelligence/pypop/blob/main/pypop7/optimizers/de/_repeat_jade.py}{data}     & YES              \\ \midrule
      CODE               & \href{https://github.com/Evolutionary-Intelligence/pypop/blob/main/pypop7/optimizers/de/_repeat_code.py}{\_repeat\_code.py}             & \href{https://github.com/Evolutionary-Intelligence/pypop/blob/main/pypop7/optimizers/de/_repeat_code.py}{data}    & YES              & TDE                & \href{https://github.com/Evolutionary-Intelligence/pypop/blob/main/pypop7/optimizers/de/_repeat_tde.py}{\_repeat\_tde.py}               & \href{https://github.com/Evolutionary-Intelligence/pypop/tree/main/docs/repeatability/tde}{figures}                & YES              \\ \midrule
      CDE                & \href{https://github.com/Evolutionary-Intelligence/pypop/blob/main/pypop7/optimizers/de/_repeat_cde.py}{\_repeat\_cde.py}               & \href{https://github.com/Evolutionary-Intelligence/pypop/blob/main/pypop7/optimizers/de/_repeat_cde.py}{data}     & YES              & CCPSO2             & \href{https://github.com/Evolutionary-Intelligence/pypop/blob/main/pypop7/optimizers/pso/_repeat_ccpso2.py}{\_repeat\_ccpso2.py}        & \href{https://github.com/Evolutionary-Intelligence/pypop/blob/main/pypop7/optimizers/pso/_repeat_ccpso2.py}{data}  & YES              \\ \midrule
      IPSO               & \href{https://github.com/Evolutionary-Intelligence/pypop/blob/main/pypop7/optimizers/pso/_repeat_ipso.py}{\_repeat\_ipso.py}            & \href{https://github.com/Evolutionary-Intelligence/pypop/blob/main/pypop7/optimizers/pso/_repeat_ipso.py}{data}   & YES              & CLPSO              & \href{https://github.com/Evolutionary-Intelligence/pypop/blob/main/pypop7/optimizers/pso/_repeat_clpso.py}{\_repeat\_clpso.py}          & \href{https://github.com/Evolutionary-Intelligence/pypop/blob/main/pypop7/optimizers/pso/_repeat_clpso.py}{data}   & YES              \\ \midrule
      CPSO               & \href{https://github.com/Evolutionary-Intelligence/pypop/blob/main/pypop7/optimizers/pso/_repeat_cpso.py}{\_repeat\_cpso.py}            & \href{https://github.com/Evolutionary-Intelligence/pypop/blob/main/pypop7/optimizers/pso/_repeat_cpso.py}{data}   & YES              & SPSOL              & \href{https://github.com/Evolutionary-Intelligence/pypop/blob/main/pypop7/optimizers/pso/_repeat_spsol.py}{\_repeat\_spsol.py}          & \href{https://github.com/Evolutionary-Intelligence/pypop/blob/main/pypop7/optimizers/pso/_repeat_spsol.py}{data}   & YES              \\ \midrule
      SPSO               & \href{https://github.com/Evolutionary-Intelligence/pypop/blob/main/pypop7/optimizers/pso/_repeat_spso.py}{\_repeat\_spso.py}            & \href{https://github.com/Evolutionary-Intelligence/pypop/blob/main/pypop7/optimizers/pso/_repeat_spso.py}{data}   & YES              & HCC                & N/A                                                                                                                                     & N/A                                                                                                                & N/A              \\ \midrule
      COCMA              & N/A                                                                                                                                     & N/A                                                                                                               & N/A              & COEA               & \href{https://github.com/Evolutionary-Intelligence/pypop/blob/main/pypop7/optimizers/cc/_repeat_coea.py}{\_repeat\_coea.py}             & \href{https://github.com/Evolutionary-Intelligence/pypop/tree/main/docs/repeatability/coea}{figures}               & YES              \\ \midrule
      COSYNE             & \href{https://github.com/Evolutionary-Intelligence/pypop/blob/main/pypop7/optimizers/cc/_repeat_cosyne.py}{\_repeat\_cosyne.py}         & \href{https://github.com/Evolutionary-Intelligence/pypop/blob/main/pypop7/optimizers/cc/_repeat_cosyne.py}{data}  & YES              & ESA                & \href{https://github.com/Evolutionary-Intelligence/pypop/blob/main/pypop7/optimizers/sa/_repeat_esa.py}{\_repeat\_esa.py}               & \href{https://github.com/Evolutionary-Intelligence/pypop/blob/main/pypop7/optimizers/sa/_repeat_esa.py}{data}      & N/A              \\ \midrule
      CSA                & \href{https://github.com/Evolutionary-Intelligence/pypop/blob/main/pypop7/optimizers/sa/_repeat_csa.py}{\_repeat\_csa.py}               & \href{https://github.com/Evolutionary-Intelligence/pypop/blob/main/pypop7/optimizers/sa/_repeat_csa.py}{data}     & YES              & NSA                & N/A                                                                                                                                     & N/A                                                                                                                & N/A              \\ \midrule
      ASGA               & \href{https://github.com/Evolutionary-Intelligence/pypop/blob/main/pypop7/optimizers/ga/_repeat_asga.py}{\_repeat\_asga.py}             & \href{https://github.com/Evolutionary-Intelligence/pypop/tree/main/docs/repeatability/asga}{data}                 & YES              & GL25               & \href{https://github.com/Evolutionary-Intelligence/pypop/blob/main/pypop7/optimizers/ga/_repeat_gl25.py}{\_repeat\_gl25.py}             & \href{https://github.com/Evolutionary-Intelligence/pypop/blob/main/pypop7/optimizers/ga/_repeat_gl25.py}{data}     & YES              \\ \midrule
      G3PCX              & \href{https://github.com/Evolutionary-Intelligence/pypop/blob/main/pypop7/optimizers/ga/_repeat_g3pcx.py}{\_repeat\_g3pcx.py}           & \href{https://github.com/Evolutionary-Intelligence/pypop/tree/main/docs/repeatability/g3pcx}{figure}              & YES              & GENITOR            & N/A                                                                                                                                     & N/A                                                                                                                & N/A              \\ \midrule
      LEP                & \href{https://github.com/Evolutionary-Intelligence/pypop/blob/main/pypop7/optimizers/ep/_repeat_lep.py}{\_repeat\_lep.py}               & \href{https://github.com/Evolutionary-Intelligence/pypop/blob/main/pypop7/optimizers/ep/_repeat_lep.py}{data}     & YES              & FEP                & \href{https://github.com/Evolutionary-Intelligence/pypop/blob/main/pypop7/optimizers/ep/_repeat_fep.py}{\_repeat\_fep.py}               & \href{https://github.com/Evolutionary-Intelligence/pypop/blob/main/pypop7/optimizers/ep/_repeat_fep.py}{data}      & NI               \\ \midrule
      CEP                & \href{https://github.com/Evolutionary-Intelligence/pypop/blob/main/pypop7/optimizers/ep/_repeat_cep.py}{\_repeat\_cep.py}               & \href{https://github.com/Evolutionary-Intelligence/pypop/blob/main/pypop7/optimizers/ep/_repeat_cep.py}{data}     & YES              & POWELL             & \href{https://github.com/Evolutionary-Intelligence/pypop/blob/main/pypop7/optimizers/ds/_repeat_powell.py}{\_repeat\_powell.py}         & \href{https://github.com/Evolutionary-Intelligence/pypop/blob/main/pypop7/optimizers/ds/_repeat_powell.py}{data}   & YES              \\ \midrule
      GPS                & N/A                                                                                                                                     & N/A                                                                                                               & N/A              & NM                 & \href{https://github.com/Evolutionary-Intelligence/pypop/blob/main/pypop7/optimizers/ds/_repeat_nm.py}{\_repeat\_nm.py}                 & \href{https://github.com/Evolutionary-Intelligence/pypop/blob/main/pypop7/optimizers/ds/_repeat_nm.py}{data}       & YES              \\ \midrule
      HJ                 & \href{https://github.com/Evolutionary-Intelligence/pypop/blob/main/pypop7/optimizers/ds/_repeat_hj.py}{\_repeat\_hj.py}                 & \href{https://github.com/Evolutionary-Intelligence/pypop/blob/main/pypop7/optimizers/ds/_repeat_hj.py}{data}      & YES              & CS                 & N/A                                                                                                                                     & N/A                                                                                                                & N/A              \\ \midrule
      BES                & \href{https://github.com/Evolutionary-Intelligence/pypop/blob/main/pypop7/optimizers/rs/_repeat_bes.py}{\_repeat\_bes.py}               & \href{https://github.com/Evolutionary-Intelligence/pypop/tree/main/docs/repeatability/bes}{figures}               & YES              & GS                 & \href{https://github.com/Evolutionary-Intelligence/pypop/blob/main/pypop7/optimizers/rs/_repeat_gs.py}{\_repeat\_gs.py}                 & \href{https://github.com/Evolutionary-Intelligence/pypop/tree/main/docs/repeatability/gs}{figures}                 & YES              \\ \midrule
      SRS                & N/A                                                                                                                                     & N/A                                                                                                               & N/A              & ARHC               & \href{https://github.com/Evolutionary-Intelligence/pypop/blob/main/pypop7/optimizers/rs/_repeat_arhc.py}{\_repeat\_arhc.py}             & \href{https://github.com/Evolutionary-Intelligence/pypop/blob/main/pypop7/optimizers/rs/_repeat_arhc.py}{data}     & YES              \\ \midrule
      RHC                & \href{https://github.com/Evolutionary-Intelligence/pypop/blob/main/pypop7/optimizers/rs/_repeat_rhc.py}{\_repeat\_rhc.py}               & \href{https://github.com/Evolutionary-Intelligence/pypop/blob/main/pypop7/optimizers/rs/_repeat_rhc.py}{data}     & YES              & PRS                & \href{https://github.com/Evolutionary-Intelligence/pypop/blob/main/pypop7/optimizers/rs/_repeat_prs.py}{\_repeat\_prs.py}               & \href{https://github.com/Evolutionary-Intelligence/pypop/tree/main/docs/repeatability/prs}{figure}                 & YES              \\ \bottomrule
    \end{tabular}
  }
  \begin{tablenotes}
    \item NI : Need to be Improved.
  \end{tablenotes}
\end{table}

\subsection{Comparisons of Computational Efficiency}

In this subsection, we will analyze the runtime efficiency (in the form of \textit{number of function evaluations}) of our implementations via empirically comparing them with those from one widely-used BBO library (called \href{https://github.com/DEAP/deap}{DEAP}). Note that \href{https://github.com/DEAP/deap}{DEAP} (which was published in 2012) mainly provided several (limited) baseline versions and has not covered the latest large-scale variants comprehensively, till now.

The test-bed is one high-dimensional (2000-d) yet light-weighted test function (named \href{https://github.com/Evolutionary-Intelligence/pypop/blob/main/pypop7/benchmarks/base_functions.py}{sphere}), since using a light-weighted test function could  make us focusing on the algorithm implementation itself rather than the external fitness function provided by the end-users. We postpone more benchmarking experiments in the following two subsections.

\begin{figure}[t]
  \centering
  \begin{minipage}{0.49\linewidth}
    \centering
    \includegraphics[width=0.9\linewidth]{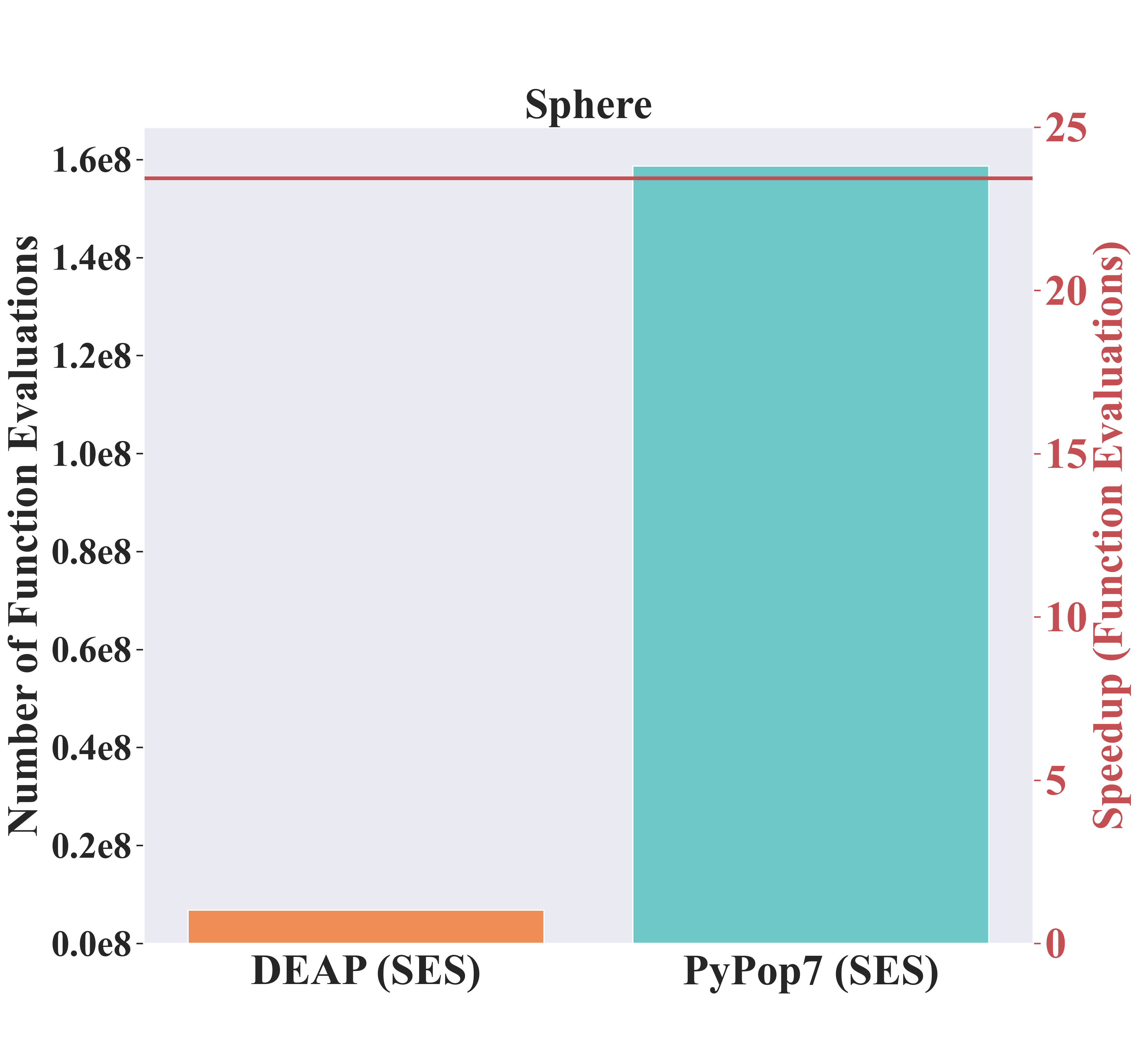}
  \end{minipage}
  \begin{minipage}{0.49\linewidth}
    \centering
    \includegraphics[width=0.9\linewidth]{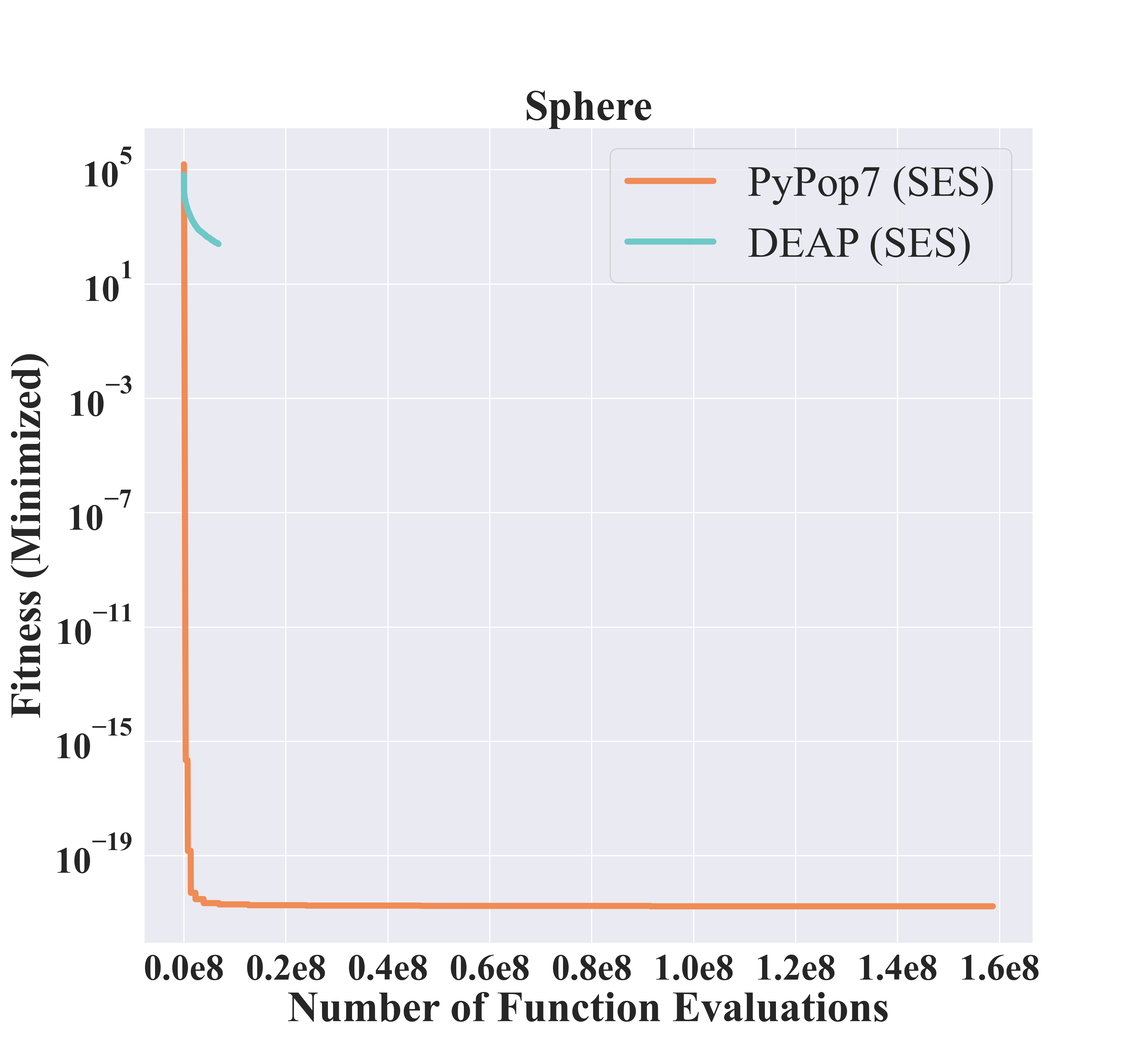}
  \end{minipage}
  \qquad
  \caption{Median comparisons of function evaluations and solution qualities of one baseline version \href{https://pypop.readthedocs.io/en/latest/es/saes.html}{SES} of \href{https://pypop.readthedocs.io/en/latest/es/es.html}{evolution strategies} from our library and the widely-used \href{https://github.com/DEAP/deap}{DEAP} library (under the same runtime for a fair comparison). Note that each of these two implementation versions is independently run 10 times on this 2000-dimensional, light-weighted test function \href{https://github.com/Evolutionary-Intelligence/pypop/blob/main/pypop7/benchmarks/base_functions.py}{\textit{sphere}}. Here we do not use the standard rotation-shift operations, different from the following computationally-expensive benchmarking process (of quadratic complexity), in order to generate light-weighted function evaluations (of only linear complexity) even in high dimensions.}
  \label{fig:median_compare_pypop7_deap_3}
\end{figure}

\begin{figure}[htbp]
  \centering
  \begin{minipage}{0.49\linewidth}
    \centering
    \includegraphics[width=0.9\linewidth]{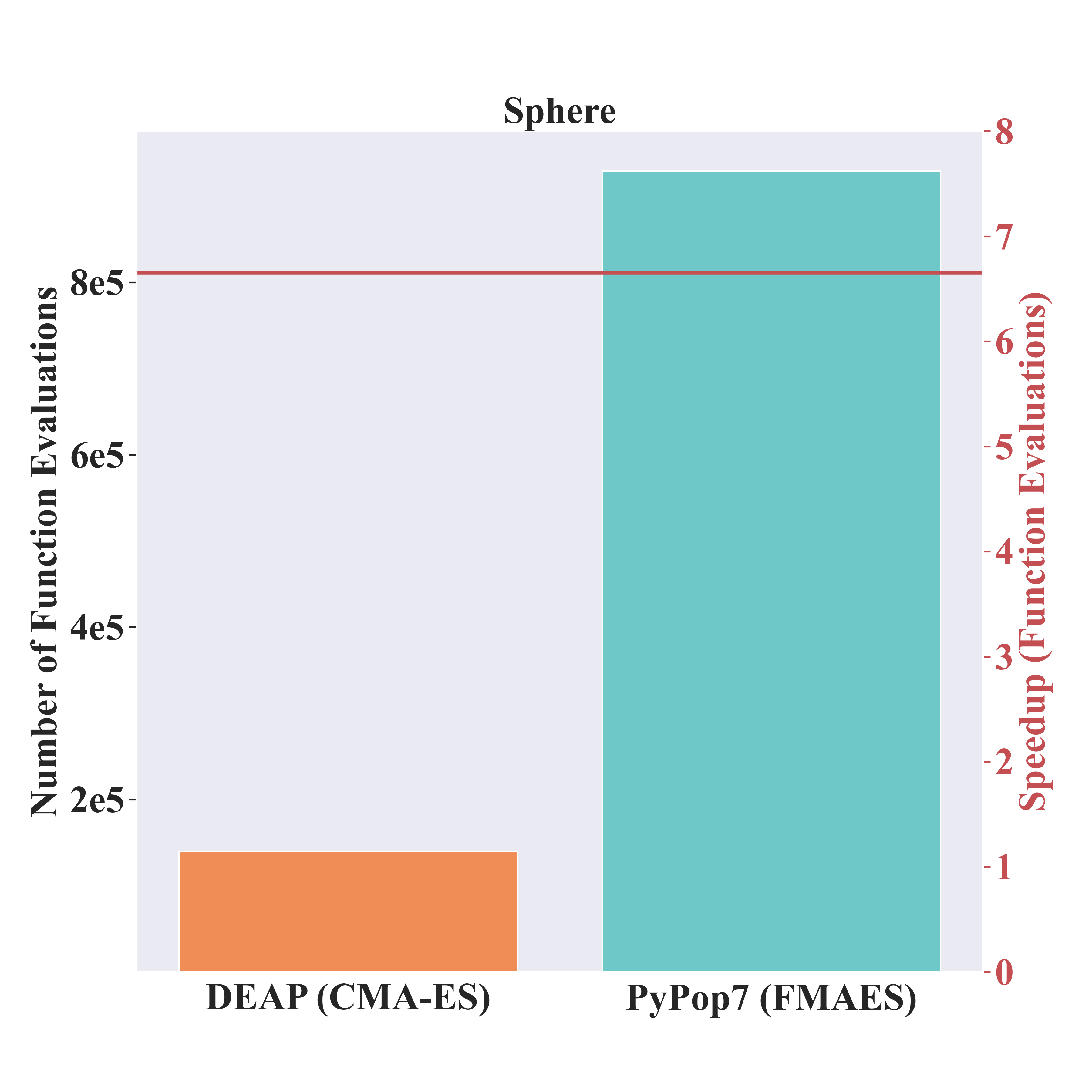}
  \end{minipage}
  \begin{minipage}{0.49\linewidth}
    \centering
    \includegraphics[width=0.9\linewidth]{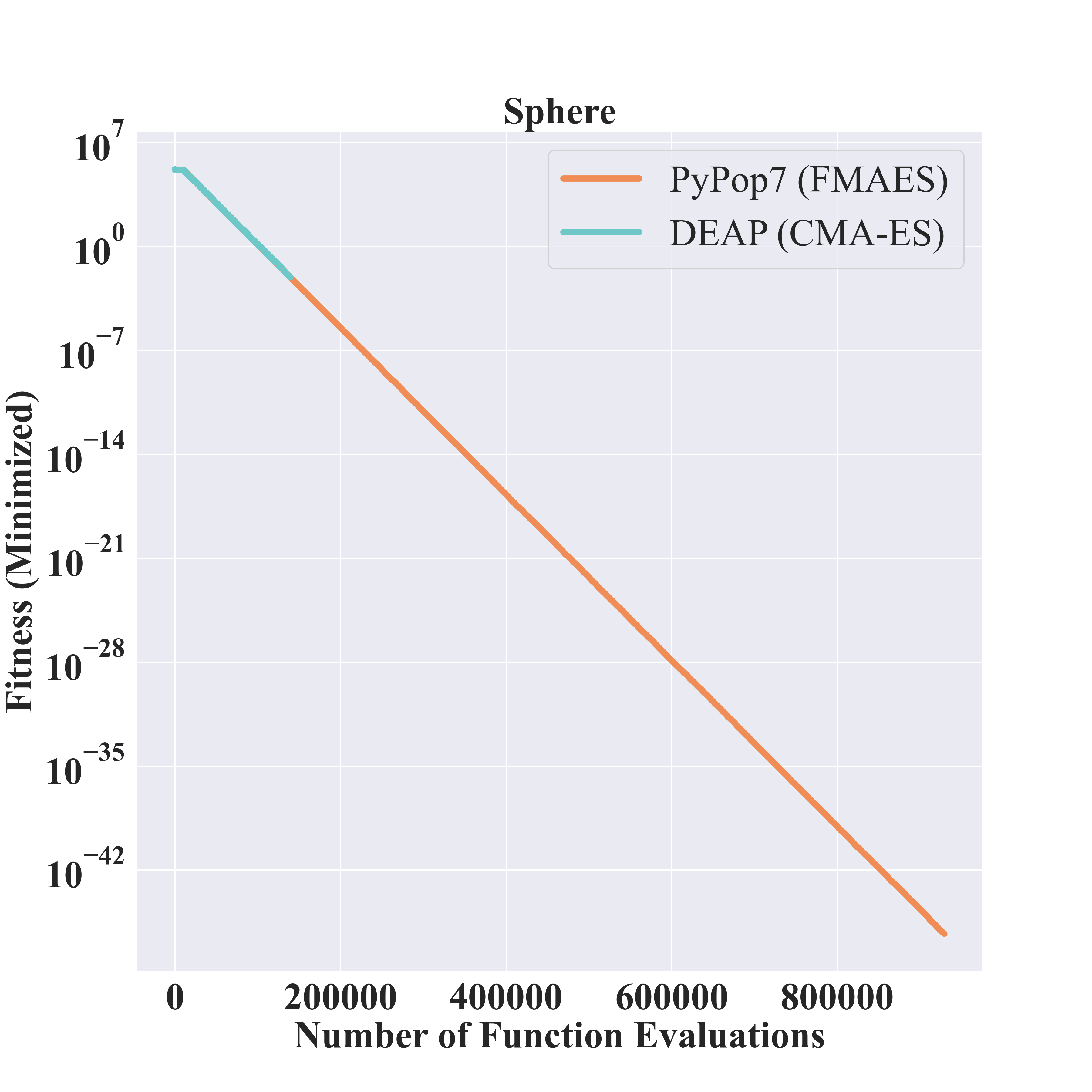}
  \end{minipage}
  \qquad

  \begin{minipage}{0.49\linewidth}
    \centering
    \includegraphics[width=0.9\linewidth]{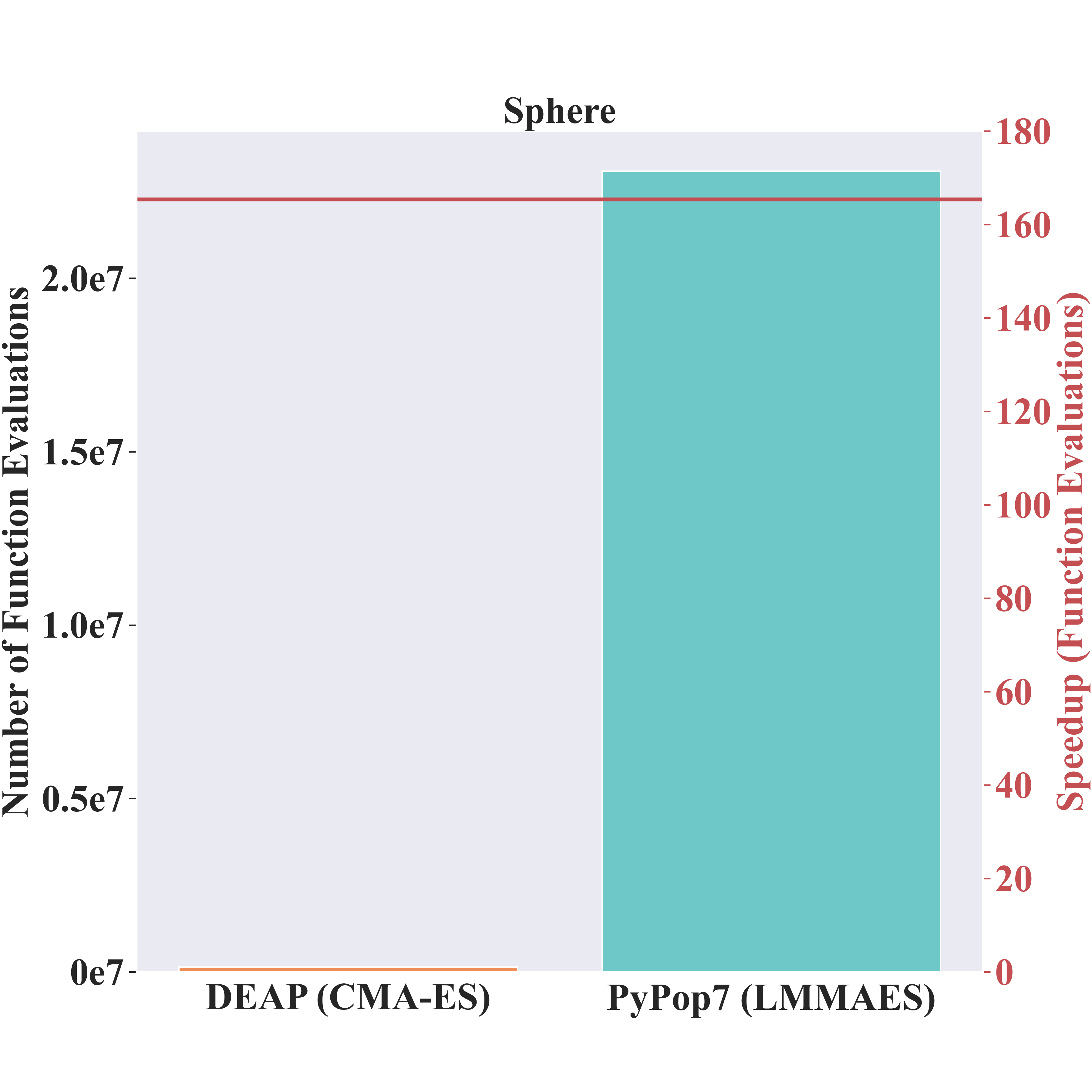}
  \end{minipage}
  \begin{minipage}{0.49\linewidth}
    \centering
    \includegraphics[width=0.9\linewidth]{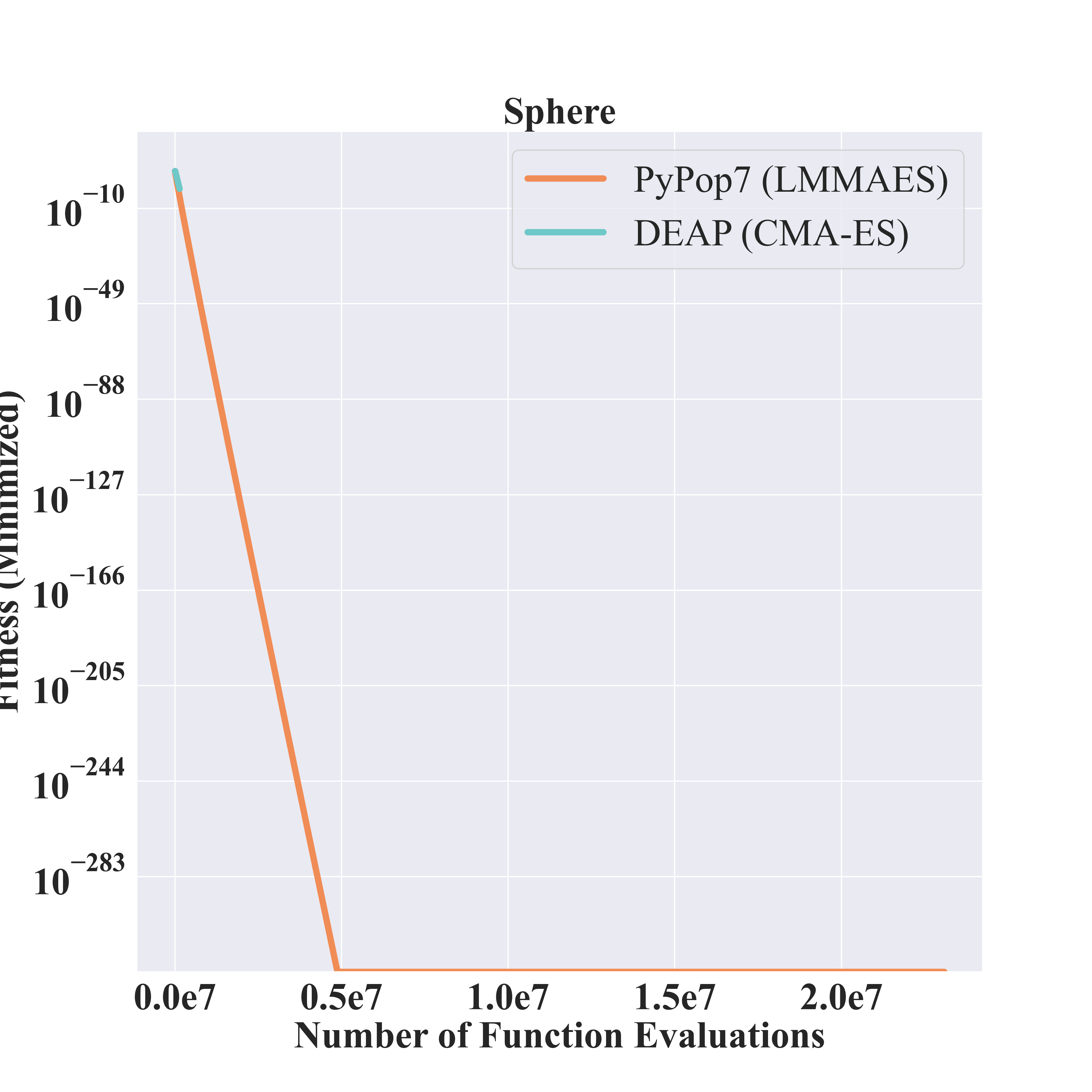}
  \end{minipage}
  \qquad

  \begin{minipage}{0.49\linewidth}
    \centering
    \includegraphics[width=0.9\linewidth]{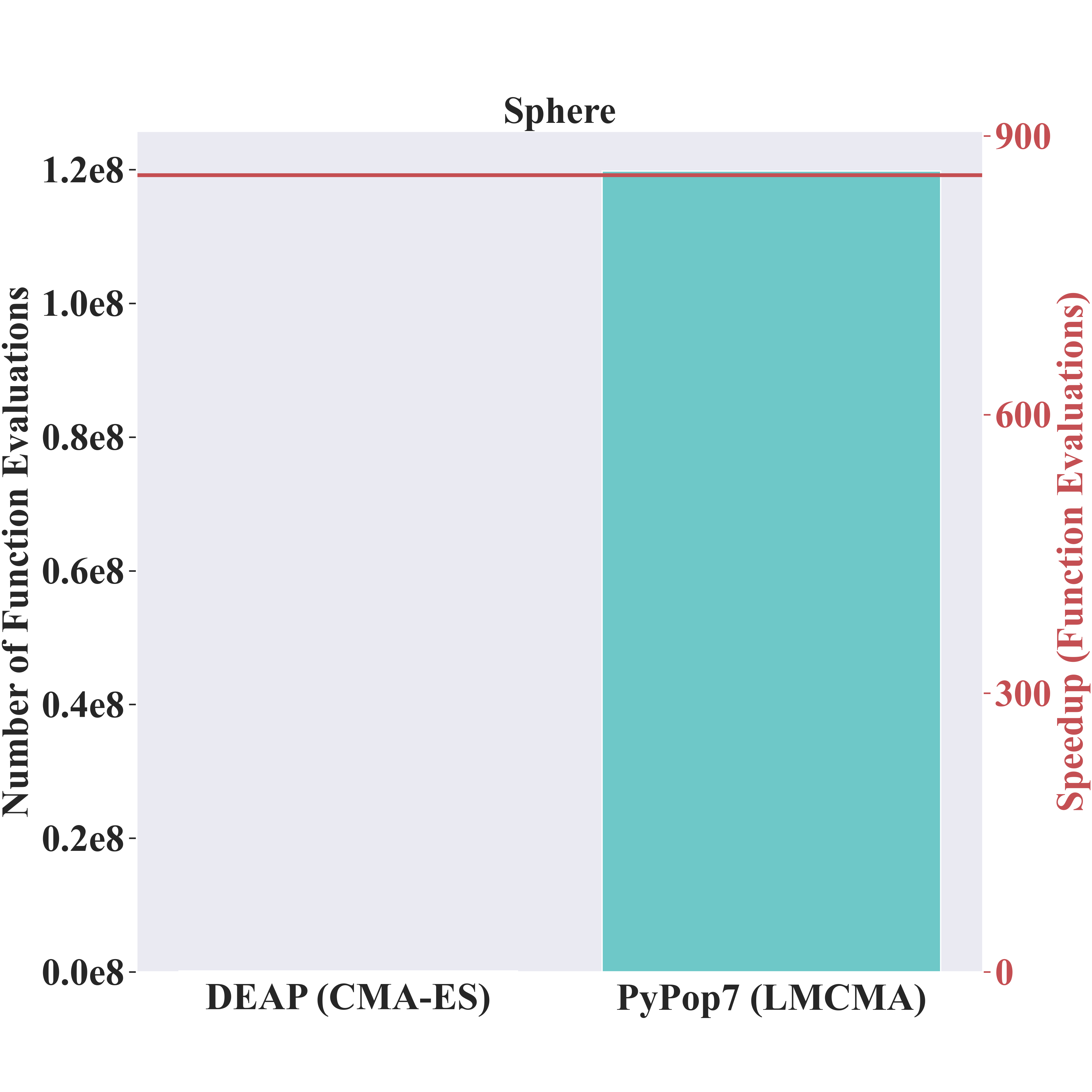}
  \end{minipage}
  \begin{minipage}{0.49\linewidth}
    \centering
    \includegraphics[width=0.9\linewidth]{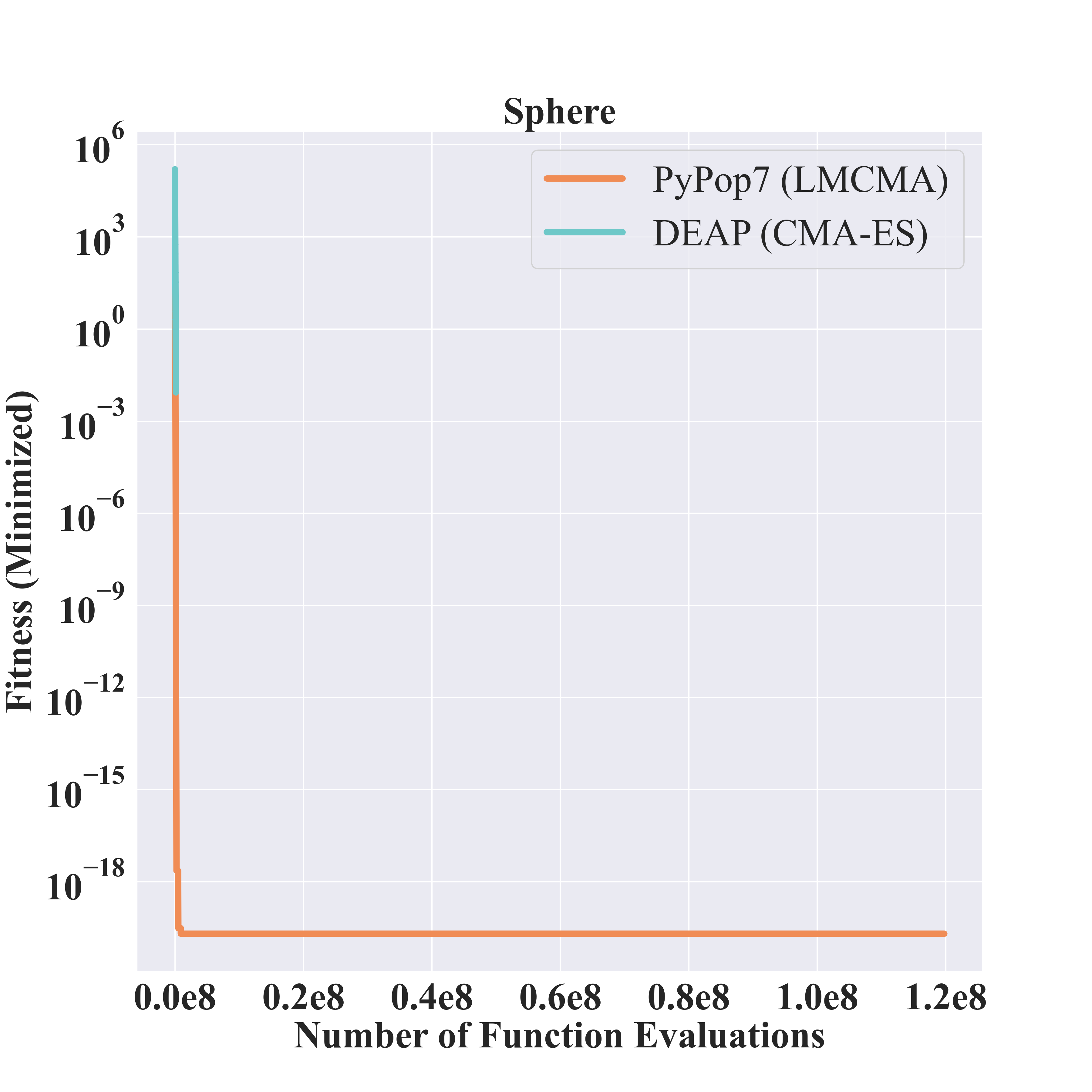}
  \end{minipage}

  \caption{Median comparisons of function evaluations and solution qualities between three large-scale ES versions of our library and \href{https://github.com/DEAP/deap}{DEAP}'s CMA-ES. The experimental settings are the same as Figure~\ref{fig:median_compare_pypop7_deap_3} (given the maximal runtime: 3 hours).}
  \label{fig:median_compare_pypop7_deap_1}
\end{figure}

\begin{figure}[htbp]
  \centering
  \begin{minipage}{0.49\linewidth}
    \centering
    \includegraphics[width=0.9\linewidth]{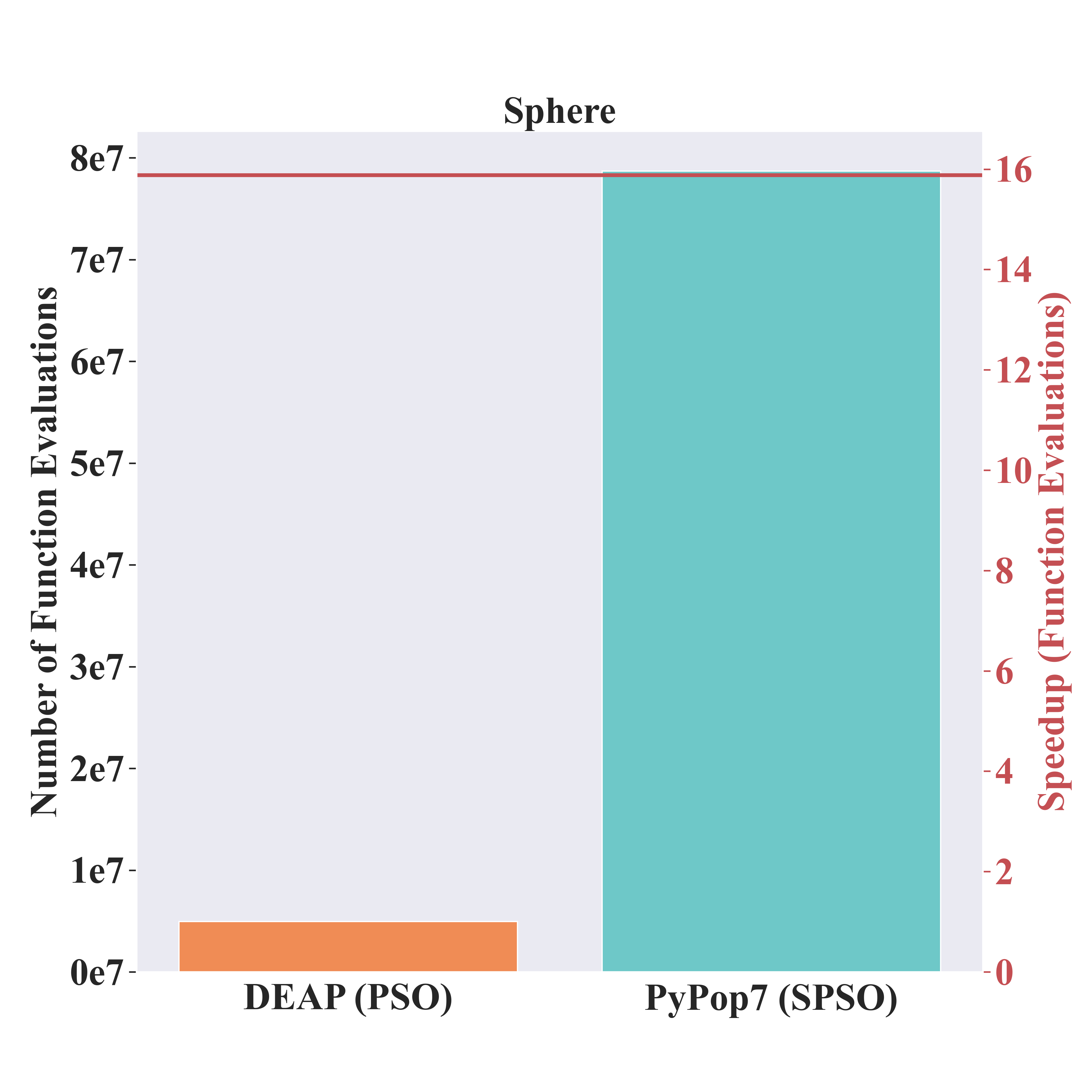}
  \end{minipage}
  \begin{minipage}{0.49\linewidth}
    \centering
    \includegraphics[width=0.9\linewidth]{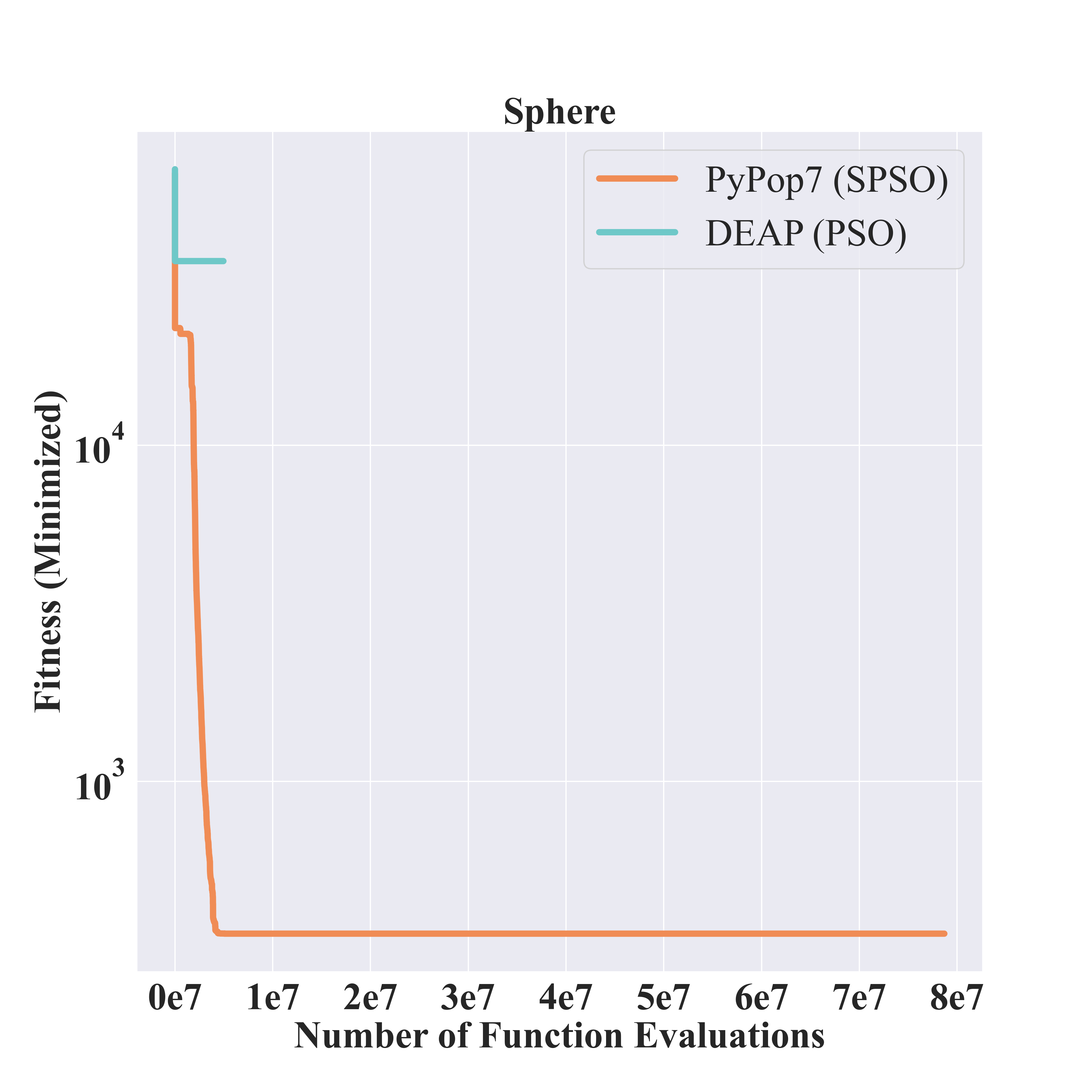}
  \end{minipage}
  \qquad

  \begin{minipage}{0.49\linewidth}
    \centering
    \includegraphics[width=0.9\linewidth]{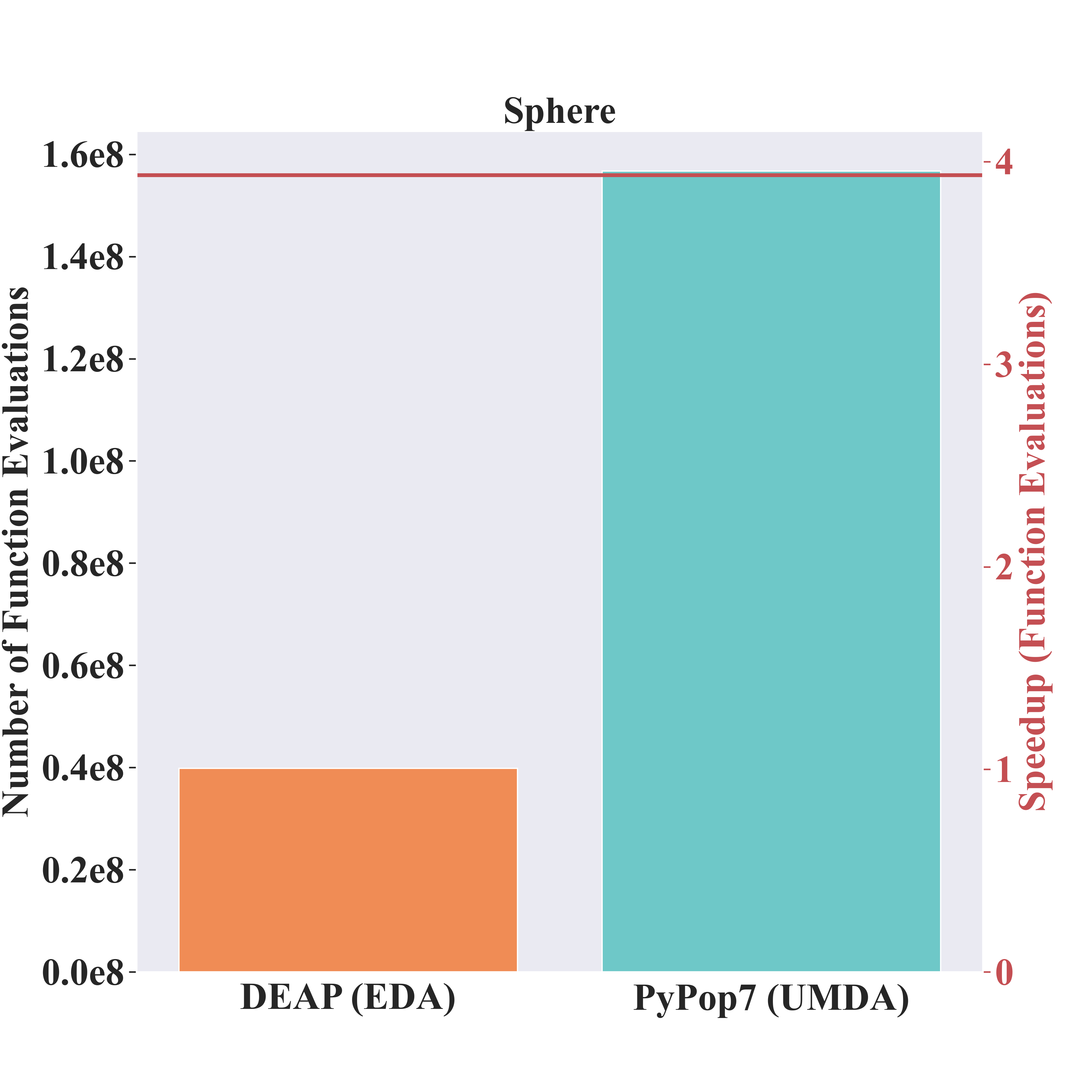}
  \end{minipage}
  \begin{minipage}{0.49\linewidth}
    \centering
    \includegraphics[width=0.9\linewidth]{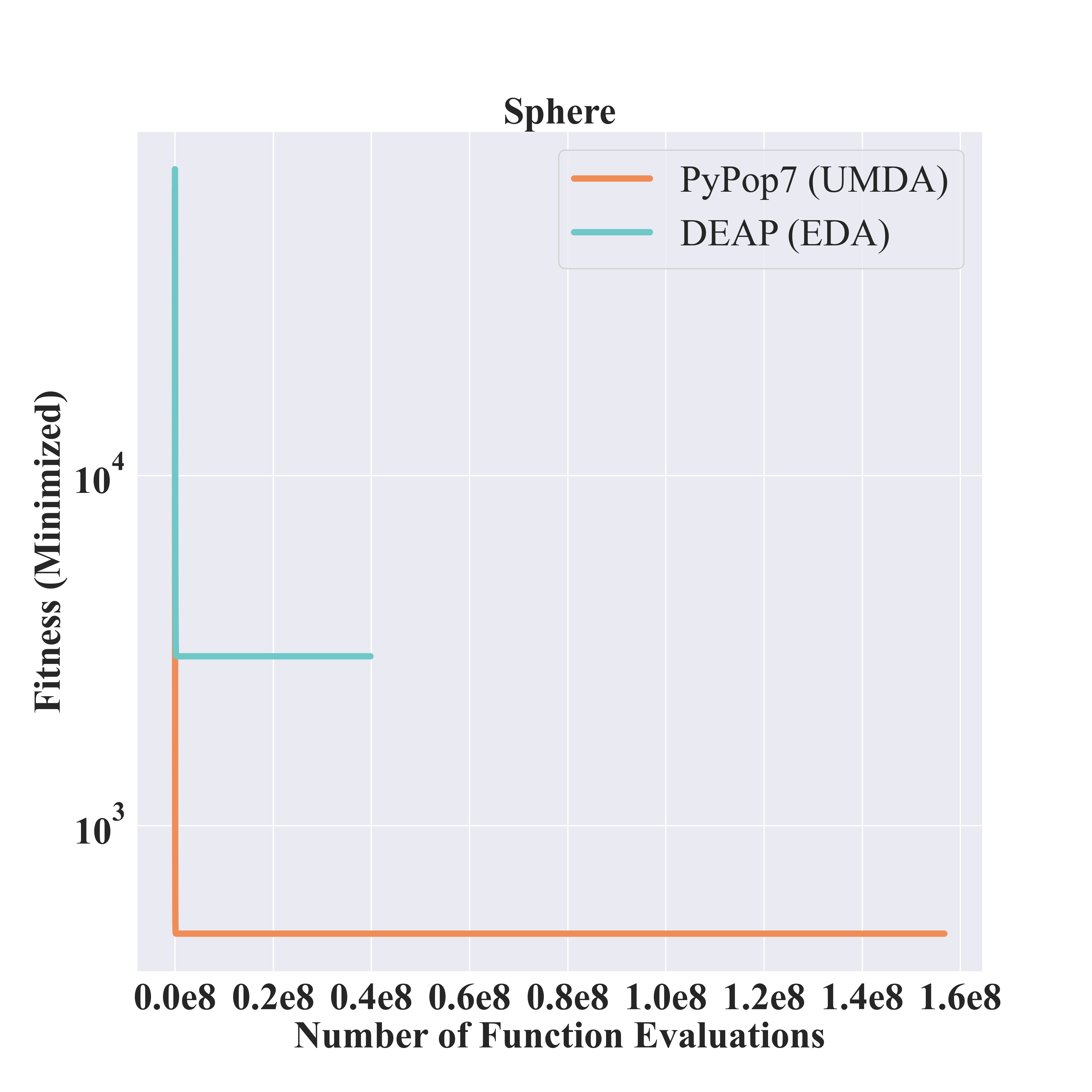}
  \end{minipage}
  \qquad

  \begin{minipage}{0.49\linewidth}
    \centering
    \includegraphics[width=0.9\linewidth]{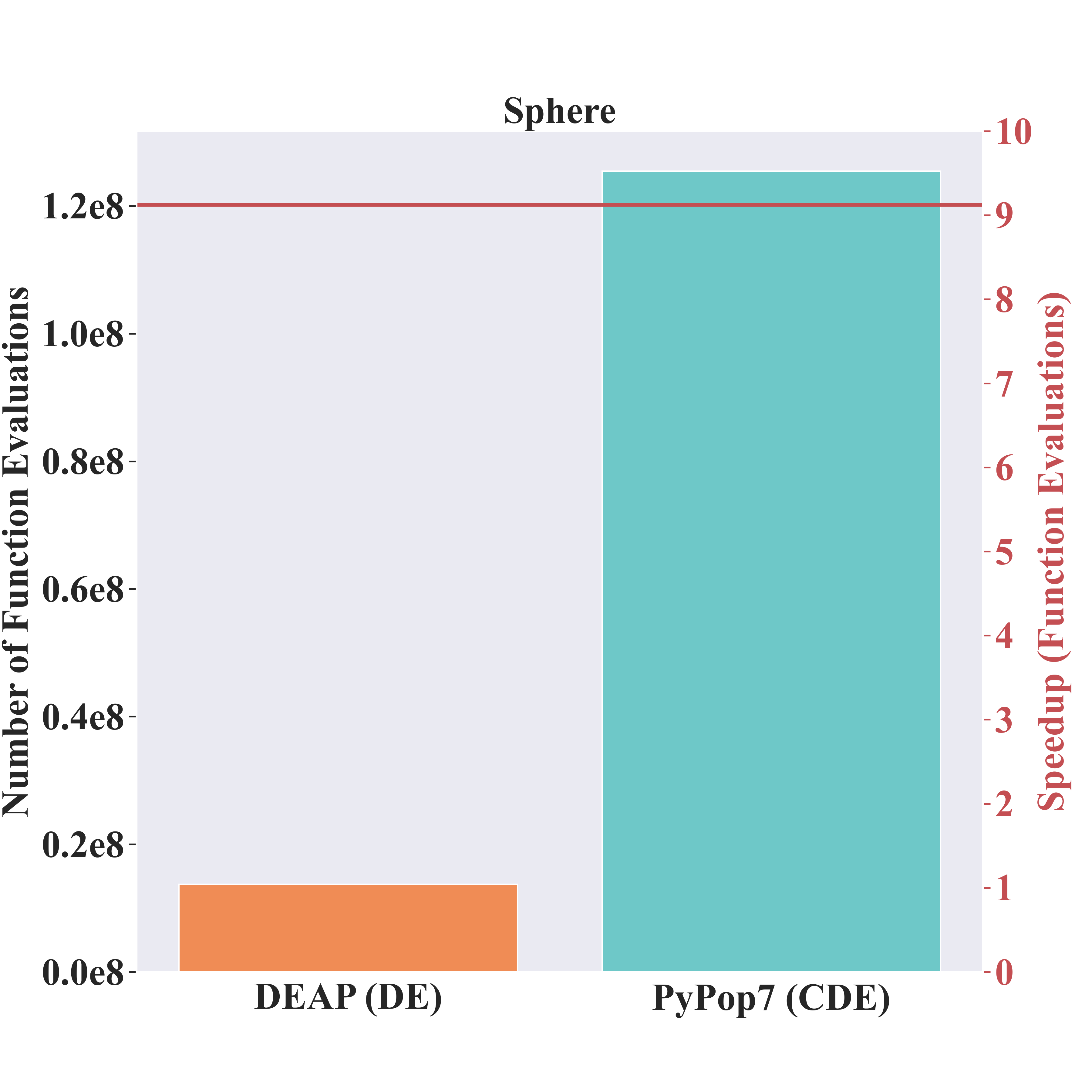}
  \end{minipage}
  \begin{minipage}{0.49\linewidth}
    \centering
    \includegraphics[width=0.9\linewidth]{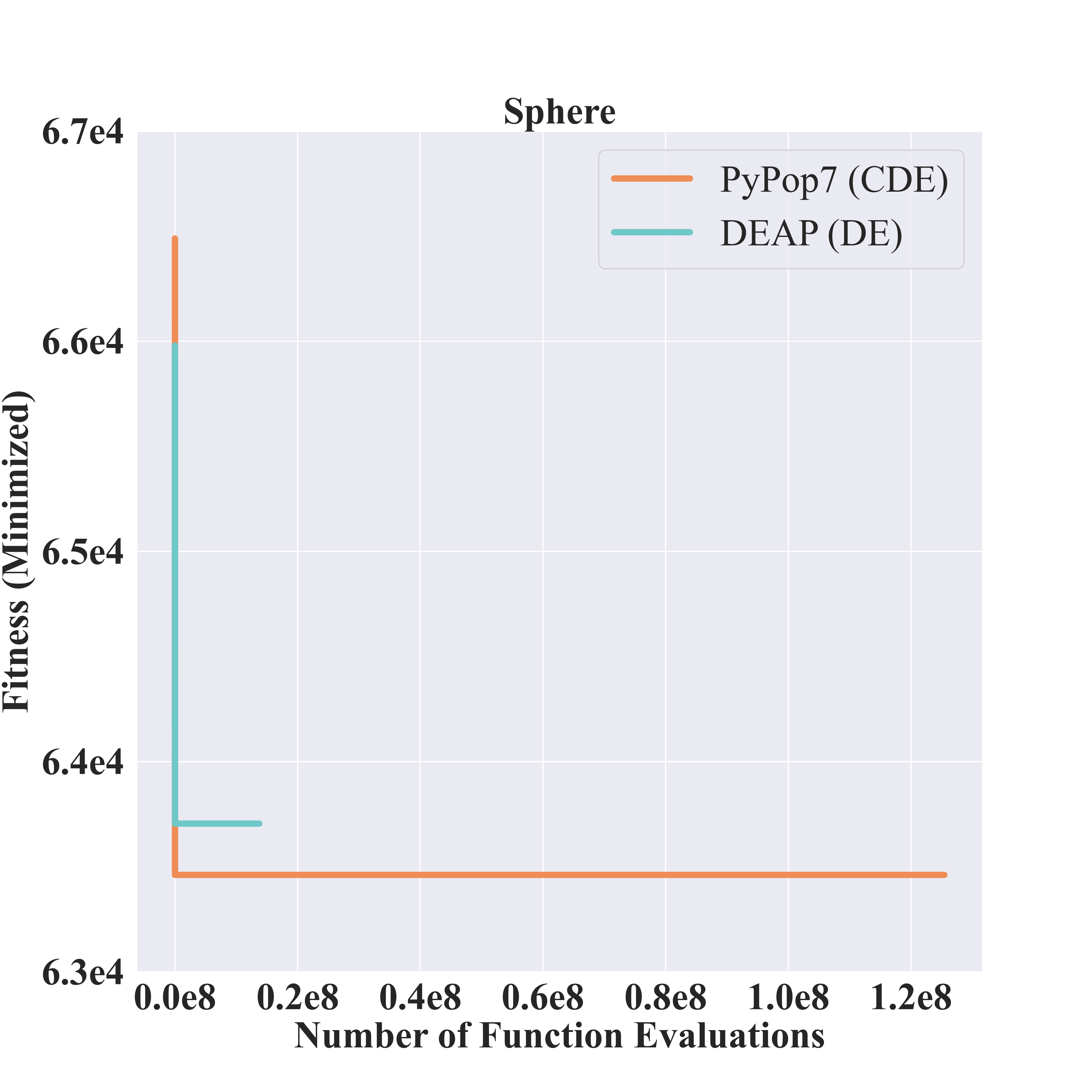}
  \end{minipage}
  \qquad
  \caption{Median comparisons of function evaluations and solution qualities of PSO, EDA, and DE between our library and the widely-used \href{https://github.com/DEAP/deap}{DEAP} library. The experimental settings are the same as Figure~\ref{fig:median_compare_pypop7_deap_3} (given the maximal runtime: 3 hours).}
  \label{fig:median_compare_pypop7_deap_2}
\end{figure}

As we can see from Figures~\ref{fig:median_compare_pypop7_deap_3},~\ref{fig:median_compare_pypop7_deap_1}, and~\ref{fig:median_compare_pypop7_deap_2}, our algorithm implementations are always better than \href{https://github.com/DEAP/deap}{DEAP}'s corresponding implementations, from both the \textit{speedup of function evaluations} and the \textit{quality of final solutions} perspectives, given the same maximal runtime (=3 hours). After carefully inspecting their own Python source code, we can conclude that different ways of storing and operating the population between two libraries (\href{https://github.com/Evolutionary-Intelligence/pypop}{PyPop7} vs. \href{https://github.com/DEAP/deap}{DEAP}) result in such a significant gap on computational efficiency. For \href{https://github.com/DEAP/deap}{DEAP} naive data types such as list are used to store and operate the population (slowly) while for \href{https://github.com/Evolutionary-Intelligence/pypop}{PyPop7} the highly-optimized data type \href{https://numpy.org/doc/stable/reference/generated/numpy.ndarray.html}{ndarray} from \href{https://numpy.org/}{\textit{NumPy}} is used as the base of population initialization and evolution, along with other high-performance scientific computing libraries such as \href{https://scipy.org/}{\textit{SciPy}}, \href{https://scikit-learn.org/stable/index.html}{\textit{Scikit-Learn}}, and \href{https://numba.pydata.org/}{\textit{Numba}}. Computational efficiency is one main goal of our open-source library, that is, developers rather than end-users are responsible for performance optimization except the customized fitness function provided by the end-user. This design practice can significantly reduce the programming and experimental overheads of end-users for large-scale BBO.

\subsection{Benchmarking on Computationally-Expensive Functions}

\begin{figure}
  \centering
  \includegraphics[width=\textwidth, height=1.2\textwidth]{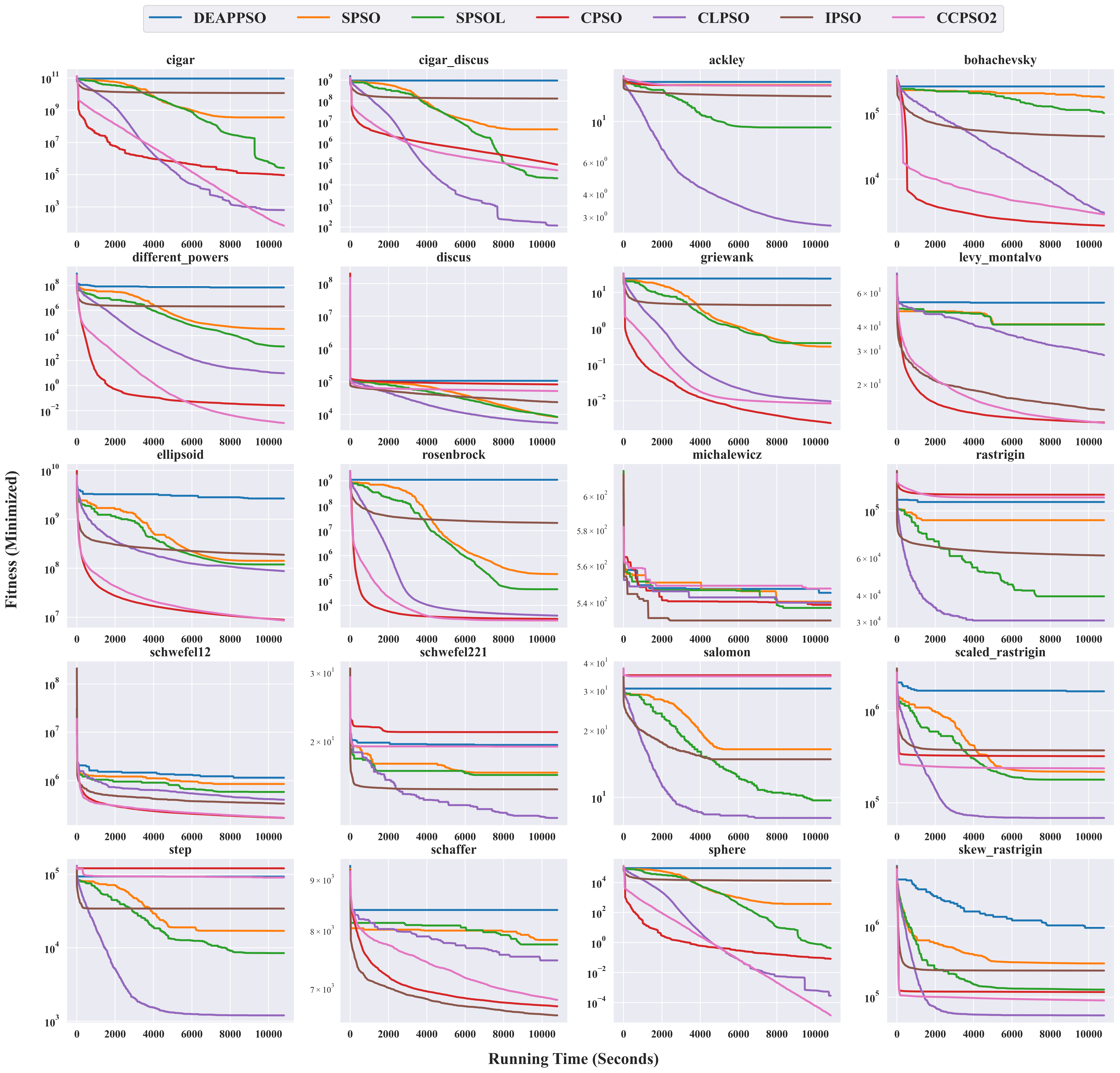}
  \caption{Median convergence rate comparisons of 7 PSO versions on 20 high-dimensional computationally-expensive test functions (with the standard rotation-and-shift operations of quadratic complexity for benchmarking).}
  \label{fig:median_compare_pso}
\end{figure}

\begin{figure}
  \centering
  \includegraphics[width=\textwidth, height=1.2\textwidth]{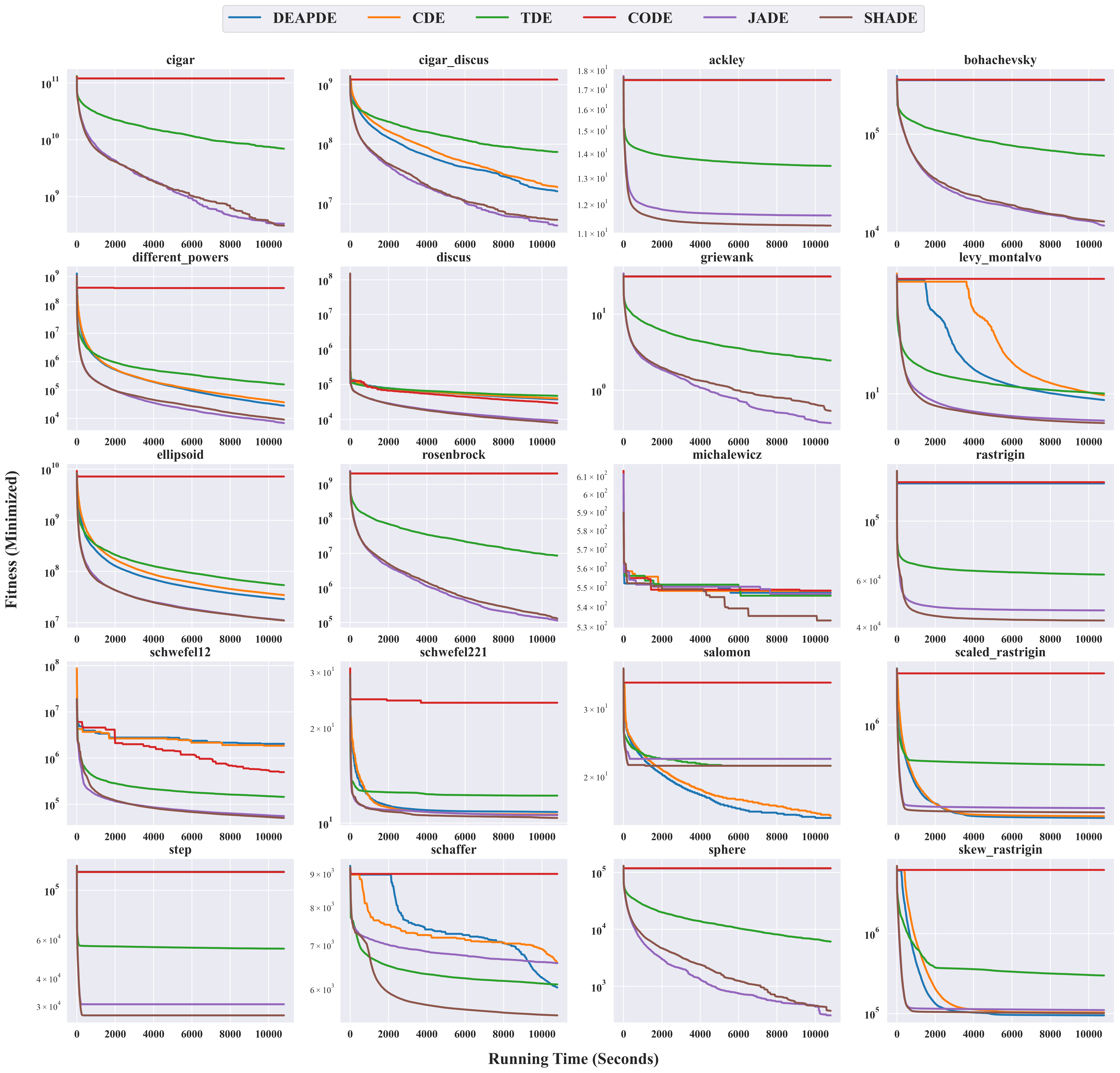}
  \caption{Median convergence rate comparisons of 6 DE versions on 20 high-dimensional computationally-expensive test functions (with the standard rotation-and-shift operations of quadratic complexity for benchmarking).}
  \label{fig:median_compare_de}
\end{figure}

\begin{figure}
  \centering
  \includegraphics[width=\textwidth, height=1.2\textwidth]{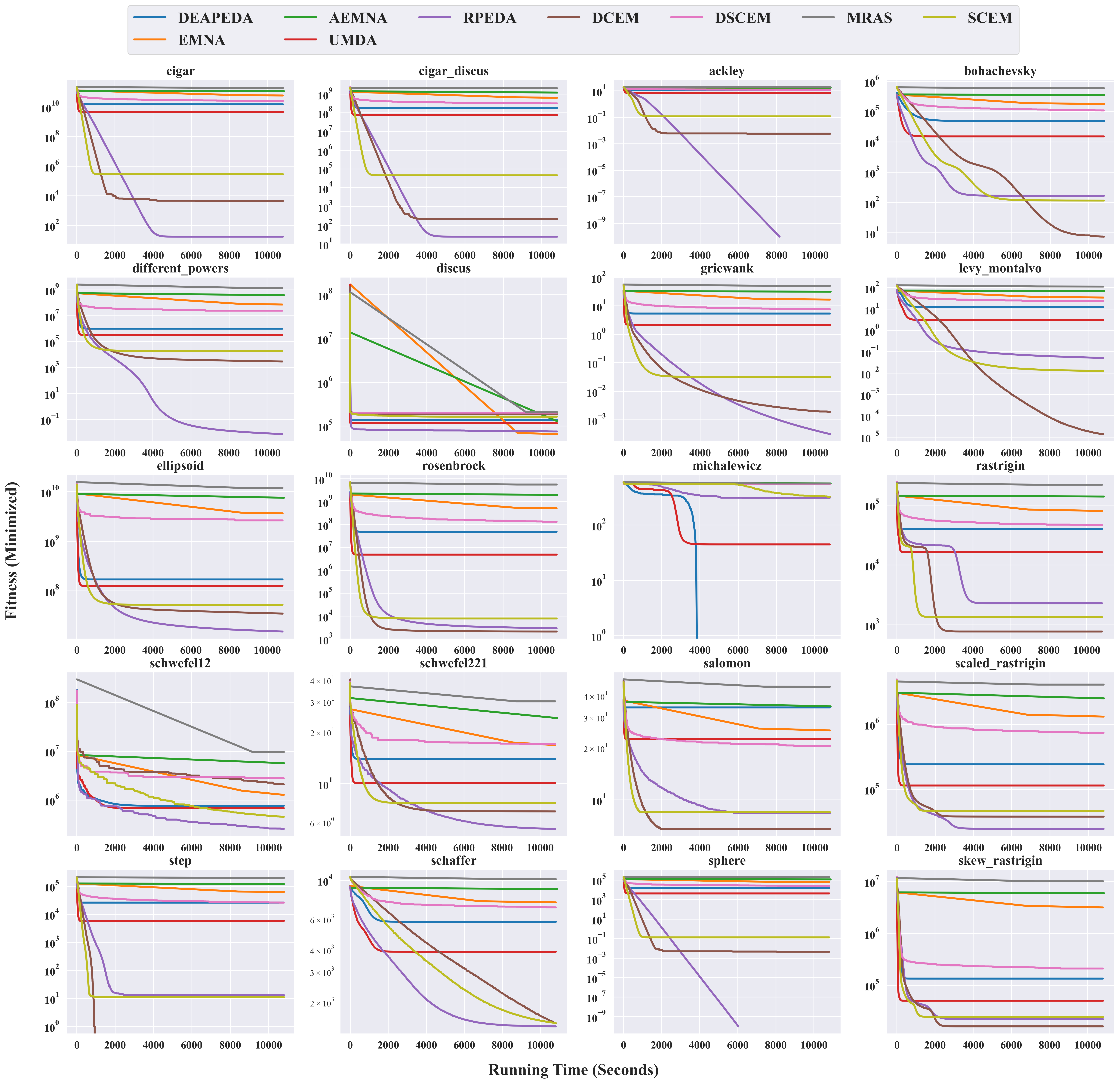}
  \caption{Median convergence rate comparisons of 9 EDA versions on 20 high-dimensional computationally-expensive test functions (with the standard rotation-and-shift operations of quadratic complexity for benchmarking).}
  \label{fig:median_compare_eda}
\end{figure}

\begin{figure}
  \centering
  \includegraphics[width=\textwidth, height=1.2\textwidth]{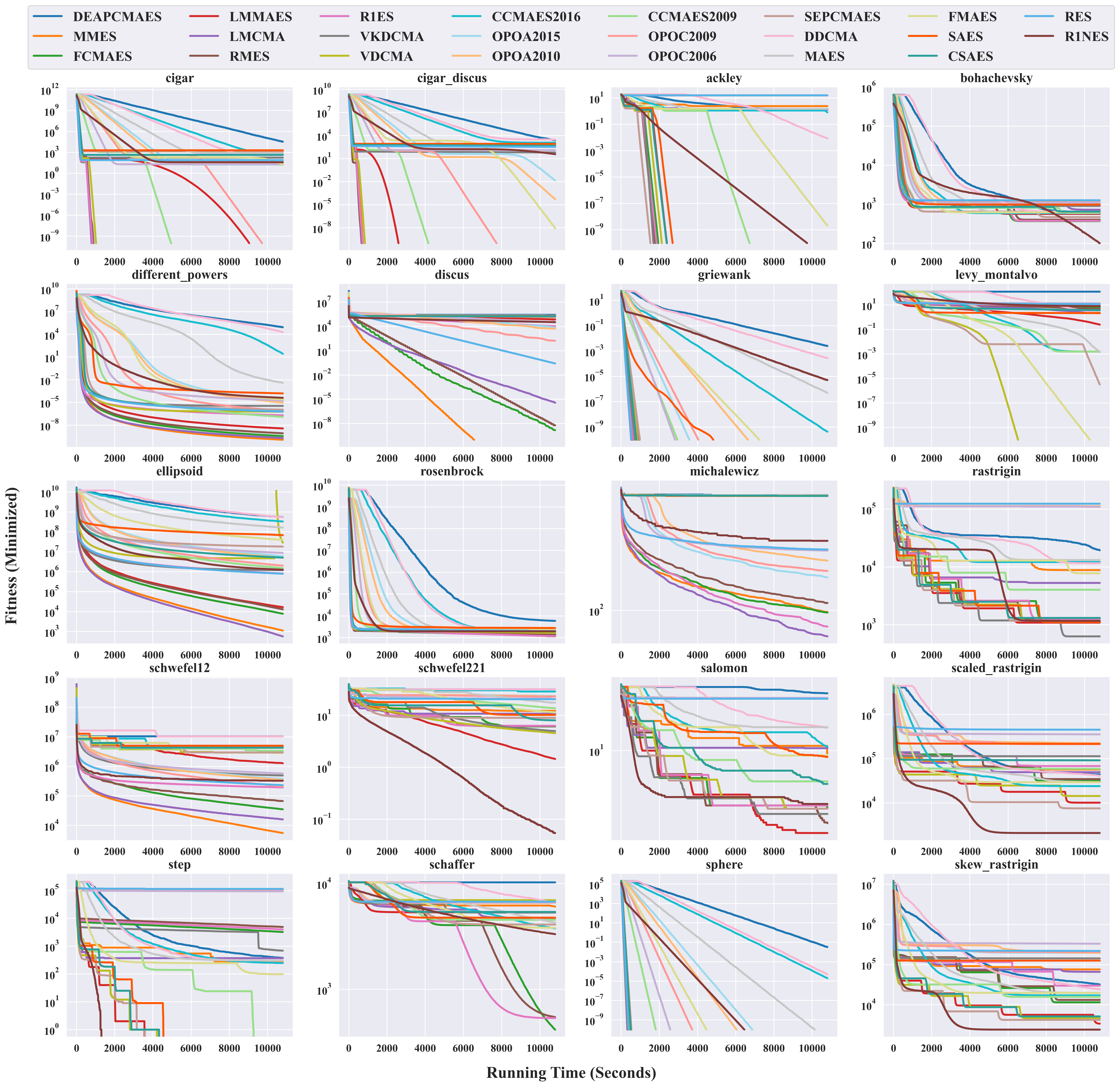}
  \caption{Median convergence rate comparisons of 23 ES versions on 20 high-dimensional computationally-expensive test functions (with the standard rotation-and-shift operations of quadratic complexity for benchmarking).}
  \label{fig:median_compare_es}
\end{figure}

To design a set of 20 computationally-expensive test functions, the standard benchmarking practice has been used here, that is, the input vector of each test functions have been rotated and shifted/transformed before fitness evaluations. For benchmarking on this large set of 2000-dimensional and computationally-expensive test functions, some of these large-scale versions from our library obtain the best solution quality on nearly all test functions under the same runtime limit (=3 hours) and the same fitness threshold (=1e-10). Please refer to Figures~\ref{fig:median_compare_pso},~\ref{fig:median_compare_de},~\ref{fig:median_compare_eda}, and~\ref{fig:median_compare_es} for detailed convergence curves of different algorithm classes on different test functions. For example, for the \href{https://github.com/Evolutionary-Intelligence/pypop/blob/main/pypop7/optimizers/pso/pso.py}{PSO} family, four large-scale variants (\href{https://github.com/Evolutionary-Intelligence/pypop/blob/main/pypop7/optimizers/pso/clpso.py}{CLPSO}, \href{https://github.com/Evolutionary-Intelligence/pypop/blob/main/pypop7/optimizers/pso/ccpso2.py}{CCPSO2}, \href{https://github.com/Evolutionary-Intelligence/pypop/blob/main/pypop7/optimizers/pso/cpso.py}{CPSO}, and \href{https://github.com/Evolutionary-Intelligence/pypop/blob/main/pypop7/optimizers/pso/ipso.py}{IPSO}) obtained the best quality of solution on 9, 6, 3, and 2 test functions, respectively.

\subsection{Benchmarking on Block-Box Classifications}

In this subsection, we choose one modern ML task (known as black-box classifications) as the base of benchmarking functions. Following currently common practices of black-box classifications, five loss functions \citep{J_MPC_bollapragada2023adaptive, J_ASTAT_li2022review, C_ICLR_ruan2020learning, C_SDM_xu2020secondorder, C_ICLR_liu2019signsgd, J_SIOPT_bollapragada2018adaptive, C_NeurIPS_liu2018zerothorder} with different landscape features are selected in our numerical experiments. Furthermore, five datasets from different fields are used for data diversity: \href{https://archive.ics.uci.edu/dataset/174/parkinsons}{\textbf{Parkinson’s disease}}, \href{https://archive.ics.uci.edu/dataset/178/semeion+handwritten+digit}{\textbf{Semeion handwritten digit}}, \href{https://archive.ics.uci.edu/dataset/233/cnae+9}{\textbf{CNAE-9}}, \href{https://archive.ics.uci.edu/dataset/171/madelon}{\textbf{Madelon}}, and \href{https://archive.ics.uci.edu/dataset/509/qsar+androgen+receptor}{\textbf{QSAR androgen receptor}}, all of which are now available at the UCI Machine Learning Repository. A combination of these 5 loss functions and 5 datasets leads to a total of 25 test functions for black-box classifications with up to $>1000$ dimensions.

\begin{figure}
  \centering
  \includegraphics[width=\textwidth, height=1.25\textwidth]{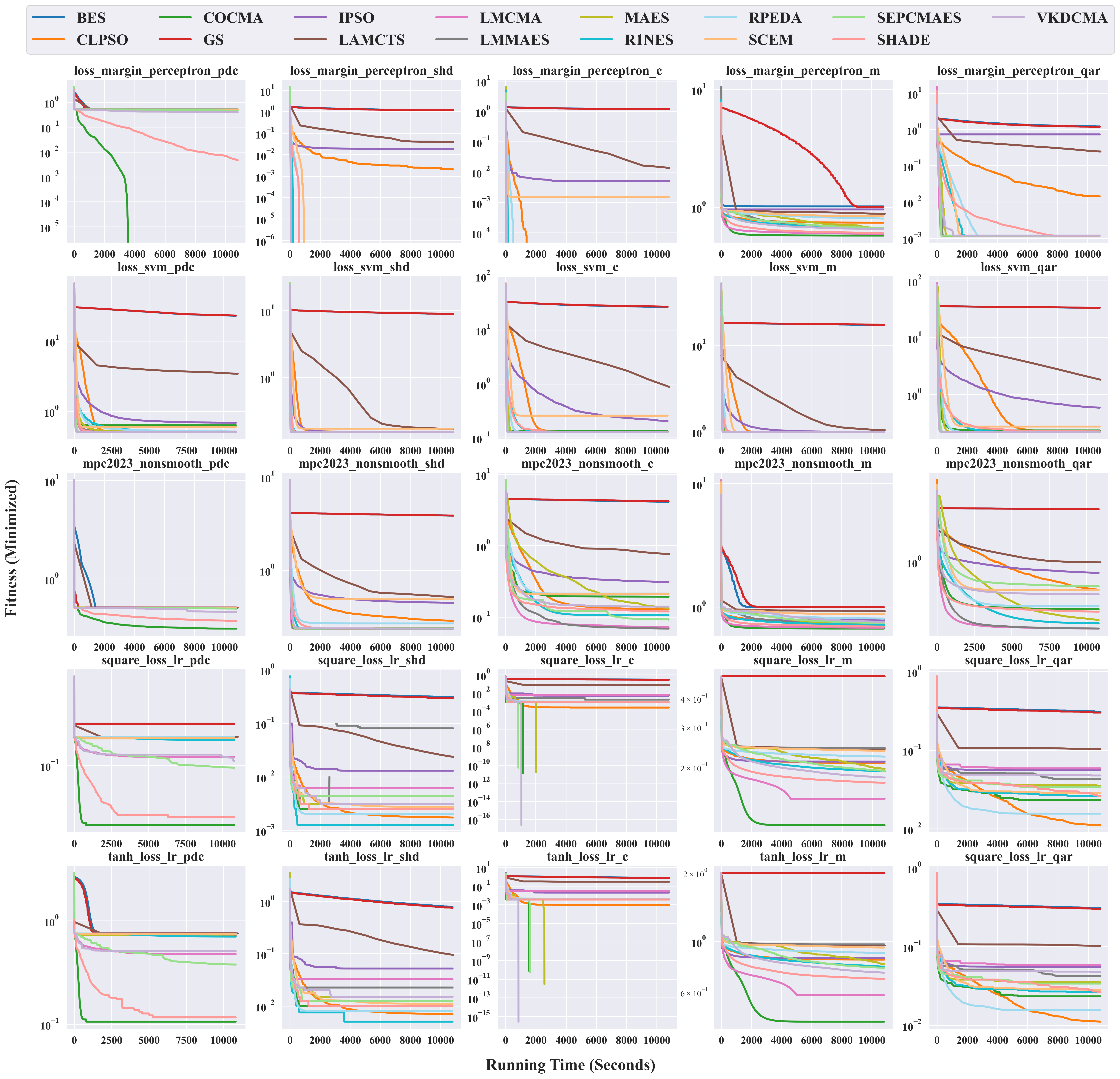}
  \caption{Comparisons of convergence curves of 15 large-scale optimizers on 25 black-box classification tasks given the maximal runtime limit (3 hours) and the fitness threshold (1e-10).}
  \label{fig:loss_functions}
\end{figure}

In our numerical experiments, we choose a total of 15 black-box optimizers from different algorithm families, each of which is independently run 14 times on every test function. The maximum of runtime to be allowed is set to 3 hours \citep{arXiv_duan2023cooperative} and the threshold of fitness is set to 1e-10 to avoid excessive accuracy optimization for all optimizers on each test function.

As is clearly shown in Figure~\ref{fig:loss_functions}, no single black-box optimizer could entirely dominate the top-ranking w.r.t. convergence curves, though some of different large-scale variants obtained the best quality of solution on different test functions. For example, COCMA \citep{J_TOMS_mei2016competitive, C_PPSN_potter1994cooperative} ranked the top on a total of 9 test functions. This may be due to that it could well exploit the sparse problem structure on these functions particularly after dataset normalization. Following it, VKDCMA \citep{C_PPSN_akimoto2016online} and CLPSO \citep{J_TEVC_liang2006comprehensive} obtained the best solution on 3 and 3 test functions, respectively. Then, each of 5 black-box optimizers (MAES \citep{J_TEVC_beyer2017simplify}, SEPCMAES \citep{C_PPSN_ros2008simple}, LMCMA \citep{J_ECJ_loshchilov2017lmcma}, LMMAES \citep{J_TEVC_loshchilov2019large}, and R1NES \citep{C_GECCOC_sun2013linear}) showed the best on 2 test functions independently. Here this ranking diversity on optimizers may empirically demonstrate the necessity to include different versions/variants of black-box optimizers in our library, seemingly in accordance with the well-established No Free Lunch Theorems (NFLT) \citep{J_TEVC_wolpert1997no}.

\section{Two Use Cases for Large-Scale BBO}

To empirically demonstrate how to properly use \href{https://github.com/Evolutionary-Intelligence/pypop}{PyPop7}, in this section we will provide two optimization examples. The first is to show its easy-to-use programming interface unified for all black-box optimizers. The following Python script shows how one large-scale ES variant called \href{https://pypop.readthedocs.io/en/latest/es/lmmaes.html}{LMMAES} \citep{J_TEVC_loshchilov2019large} minimizes the popular \href{https://github.com/Evolutionary-Intelligence/pypop/blob/main/pypop7/benchmarks/base_functions.py}{Rosenbrock} test function \citep{J_ECJ_kok2009locating}.

\begin{lstlisting}[style = python]
>>> import numpy as np
>>> from pypop7.benchmarks.base_functions import rosenbrock  # notorious test function
>>> ndim_problem = 1000 # dimension of fitness (cost) function to be minimized
>>> problem = {"fitness_function": rosenbrock, # fitness function to be minimized
...              "ndim_problem": ndim_problem, # function dimension
...              "lower_boundary": -5.0*np.ones((ndim_problem,)), # lower search boundary
...              "upper_boundary": 5.0*np.ones((ndim_problem,))}  # upper search boundary
>>> from pypop7.optimizers.es.lmmaes import LMMAES # or using any other optimizers
>>> options = {"fitness_threshold": 1e-10, # fitness threshold to terminate evolution
...              "max_runtime": 3600, # to terminate evolution when runtime exceeds 1 hour
...              "seed_rng": 0, # seed of random number generation for repeatability
...              "x": 4.0*np.ones((ndim_problem,)), # initial mean of search distribution
...              "sigma": 3.0, # initial global step-size of search distribution
...              "verbose": 500}  # to print verbose information every 500 generations
>>> lmmaes = LMMAES(problem, options) # to initialize this black-box optimizer
>>> results = lmmaes.optimize() # to run its time-consuming search process on high dimensions
>>> # to print the best-so-far fitness found and the number of function evaluations used
>>> print(results["best_so_far_y"], results["n_function_evaluations"])
\end{lstlisting}

The second is to present the benchmarking process of one black-box optimizer on the well-documented \href{https://github.com/numbbo/coco}{COCO}/BBOB platform \citep{J_ASOC_varelas2020benchmarking}, which is shown below.

\begin{lstlisting}[style = python]
>>> import os
>>> import webbrowser  # for post-processing in the browser
>>> import numpy as np
>>> import cocoex  # experimentation module of `COCO'
>>> import cocopp  # post-processing module of `COCO'
>>> from pypop7.optimizers.es.maes import MAES
>>> suite, output = "bbob", "COCO-PyPop7-MAES"
>>> budget_multiplier = 1e3  # or 1e4, 1e5, ...
>>> observer = cocoex.Observer(suite, "result_folder:" + output)
>>> minimal_print = cocoex.utilities.MiniPrint()
>>> for function in cocoex.Suite(suite, "", ""):
...     function.observe_with(observer)  # to generate data for `cocopp' post-processing
...     sigma = np.min(function.upper_bounds - function.lower_bounds) / 3.0
...     problem = {"fitness_function": function,
...                "ndim_problem": function.dimension,
...                "lower_boundary": function.lower_bounds,
...                "upper_boundary": function.upper_bounds}
...     options = {"max_function_evaluations": function.dimension * budget_multiplier,
...                "seed_rng": 2022,
...                "x": function.initial_solution,
...                "sigma": sigma}
...     solver = MAES(problem, options)
...     print(solver.optimize())
>>> cocopp.main(observer.result_folder)
>>> webbrowser.open("file://" + os.getcwd() + "/ppdata/index.html")
\end{lstlisting}

For more examples, please refer to its online documentations: \href{https://pypop.rtfd.io/}{pypop.rtfd.io}. Note that we have provided at least one example for each black-box optimizer in its corresponding API online document.

\section{Conclusion}

In this paper, we have provided an open-source pure-Python library (called \href{https://github.com/Evolutionary-Intelligence/pypop}{PyPop7}) for BBO with modular coding structures and full-fledged online documentations. Up to now, this light-weighted library has been used not only by our own work, e.g., \citep{C_PPSN_duan2022collective} and \citep{arXiv_duan2023cooperative}, but also by other work, such as, prompt tuning of vision-language models \citep{C_IJCAI_yu2023blackbox}, nonlinear optimization for radiotherapy\footnote{\url{https://github.com/pyanno4rt/pyanno4rt}}, and robotics planning/control \citep{arXiv_zhang2024invariant, arXiv_lee2023planner}. Please refer to its \href{https://pypop.rtfd.io/}{online} documentations for an up-to-date summary of its applications.

As next steps, we plan to further enhance its capability of BBO from five aspects, as shown in the following:

\begin{itemize}
  \item Massive parallelism \citep{J_JMLR_chalumeau2024qdax, C_GECCO_lange2023evosax},
  \item Constrains handling \citep{arXiv_hellwig2024analyzing},
  \item Noisy optimization \citep{J_MLST_hase2021olympus, J_TEVC_hansen2009method, J_CMAME_beyer2000evolutionary},
  \item Meta-learning/optimization \citep{C_ICLR_lange2023discovering, C_ICML_vicol2023lowvariance, C_NeurIPS_li2023variancereduced, C_ICML_vicol2021unbiased}, and
  \item Automatic algorithm design, in particular automated algorithm selection/configuration \citep{J_JAIR_schede2022survey, J_ECJ_kerschke2019automated}.
\end{itemize}

\acks{This work is supported by the Guangdong Basic and Applied Basic Research Foundation under Grants No. 2024A1515012241 and 2021A1515110024, the Shenzhen Fundamental Research Program under Grant No. JCYJ20200109141235597, and the Program for Guangdong Introducing Innovative and Entrepreneurial Teams under Grant No. 2017ZT07X386.}

\bibliography{main}

\end{document}